\documentclass[10pt,twocolumn,letterpaper]{article}

\usepackage[pagenumbers]{cvpr} 

\usepackage{graphicx}
\usepackage{amsmath}
\usepackage{amssymb}
\usepackage[table]{xcolor}
\usepackage{booktabs}
\usepackage{multirow}
\usepackage{pifont}
\usepackage{xcolor}
\usepackage{siunitx}
\usepackage{algpseudocode} 
\usepackage{algorithm} 
\usepackage{cuted}
\usepackage{capt-of}
\usepackage{stmaryrd}
\usepackage{soul}
\usepackage{commath}
\usepackage{enumitem}
\usepackage{xfrac} 
\usepackage{flushend}
\usepackage[accsupp]{axessibility}
\usepackage[pagebackref=true,breaklinks=true,colorlinks,citecolor=blue,bookmarks=false]{hyperref}

\usepackage[capitalize]{cleveref}
\crefname{section}{Sec.}{Secs.}
\Crefname{section}{Section}{Sections}
\Crefname{table}{Table}{Tables}
\crefname{table}{Tab.}{Tabs.}

\def\cvprPaperID{5927} 
\def\confName{CVPR}
\def\confYear{2022}

\definecolor{dgreen}{rgb}{0.0, 0.5, 0.0}
\definecolor{better}{rgb}{0.19, 0.55, 0.91}
\definecolor{worse}{rgb}{0.82, 0.1, 0.26}
\definecolor{lightgray}{gray}{0.9}
\newcommand{\cmark}{\textcolor{better}{\ding{51}}}%
\newcommand{\xmark}{\textcolor{worse}{\ding{55}}}%

\newcommand{\Tx}{N_{\text{Tx}}}
\newcommand{\Rx}{N_{\text{Rx}}}
\newcommand{\BinA}{B_{\text{A}}}
\newcommand{\BinR}{B_{\text{R}}}
\newcommand{\BinD}{B_{\text{D}}}
\newcommand{\BinE}{B_{\text{E}}}

\newcommand{\dataset}{RADIal}

\newcommand{\method}{FFT-RadNet}

\newcommand{\vx}{\mathbf{x}}
\newcommand{\vy}{\mathbf{y}}

\newcommand{\parag}[1]{\smallskip\noindent\textbf{#1}~}
\newcommand{\paragnoskip}[1]{\noindent\textbf{#1}~}

\makeatletter
\DeclareRobustCommand\onedot{\futurelet\@let@token\@onedot}
\def\@onedot{\ifx\@let@token.\else.\null\fi\xspace}

\def\ie{\emph{i.e}\onedot}

\makeatother

\DeclareUnicodeCharacter{2212}{-}

\begin{document}

\title{Raw High-Definition Radar for Multi-Task Learning}
\author{Julien Rebut$^1$
\and
Arthur Ouaknine$^{1,2}$
\and
Waqas Malik$^3$
\and
Patrick Pérez$^1$
\and {\small 1:\,Valeo.ai, Paris, France\quad 2:\,T\'el\'ecom\,Paris, Paris, France\quad 3:\,Valeo North America Inc., San Mateo, CA}}
\maketitle

\begin{strip}\centering
\includegraphics[width=1.0\textwidth]{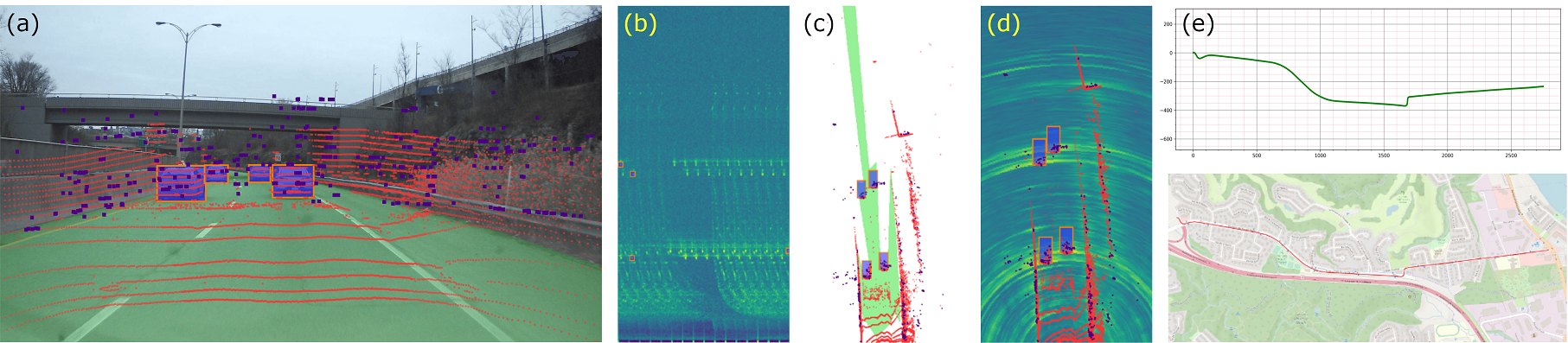}
\captionof{figure}{\textbf{Overview of our \dataset~dataset}. \dataset~includes a set of 3 sensors (camera, laser scanner, high-definition radar) and comes with GPS and vehicle's CAN traces; 25k synchronized samples are recorded is raw format. (a) Camera image with projected laser point cloud in red and radar point cloud in indigo, vehicle annotation in orange and free-driving-space annotation in green; (b) Radar power spectrum with bounding box annotations; (c) Free-driving-space annotation in bird-eye view, with vehicles annotated with orange bounding boxes, radar point cloud in indigo and laser point cloud in red; (d) Range-azimuth map in Cartesian coordinates overlayed with radar point cloud and laser point cloud; (e) GPS trace in red and odometry trajectory reconstruction in green.
}

\label{fig:teaser}
\end{strip}

\begin{abstract}
With their robustness to adverse weather conditions and ability to measure speeds, radar sensors have been part of the automotive landscape for more than two decades. 
Recent progress toward High Definition (HD) Imaging radar has driven the 
angular resolution below the degree, thus approaching 
laser scanning performance. 
However, the amount of data a HD radar delivers and the computational cost to estimate the angular positions remain a challenge. 
In this paper, we propose a novel HD radar sensing model, \emph{\method}, that eliminates the overhead of computing the range-azimuth-Doppler 3D tensor, learning instead to recover angles from a range-Doppler spectrum. \method~is trained both to detect vehicles and to segment free driving space. On both tasks, it competes with the most recent radar-based models while requiring less compute and memory. Also, we collected and annotated 2-hour worth of raw data from synchronized automotive-grade sensors (camera, laser, HD radar) in various environments (city street, highway, countryside road). This unique dataset, nick-named \emph{\dataset}~for ``Radar, LiDAR \emph{et al}.'', is available at \url{https://github.com/valeoai/RADIal}.
\end{abstract}

\section{Introduction}
\label{Introduction}

\begin{table*}[t]
\renewcommand{\tabcolsep}{1.3em}
\rowcolors{3}{}{lightgray}
    \centering
    \resizebox{0.9\linewidth}{!}{\begin{tabular}{lccccccccccc}
    \toprule
    \multirow{5}{*}{Dataset} & \multirow{5}{*}{Year} &  \multirow{5}{*}{Scale} & \multicolumn{5}{c}{Radar data} &  \multirow{5}{*}{\rotatebox{90}{Radar type}} &  \multirow{5}{*}{\rotatebox{90}{Modalities}} & \multirow{5}{*}{\rotatebox{90}{Sequence}} & \multirow{5}{*}{Annot. type}  \\ \cmidrule{4-8}
    & & & \rotatebox{90}{ADC} & \rotatebox{90}{RAD} & \rotatebox{90}{RA or RD} & \rotatebox{90}{PC} & \rotatebox{90}{Doppler}  &  &  &  \\
    \midrule
    nuScenes \cite{nuscenes_2020} & 2019 & large & \xmark & \xmark & \xmark & \cmark & \cmark & LD &   CLO & \cmark  & 3D boxes \\
    Astyx \cite{automotive_Meyer_2019} & 2019 & small & \xmark & \xmark & \xmark & \cmark & \cmark & HD & CL  & \xmark  & 3D boxes \\
    RadarRobotCar \cite{barnes_oxford_2020} & 2020 &  large  & \xmark & \xmark & \cmark & \xmark & \xmark & S & CLO  &  \cmark & \xmark  \\
    CARRADA \cite{carrada_ouaknine_2021} & 2020 &  small  & \xmark & \cmark & \cmark & \cmark & \cmark & LD & C  & \cmark  &  2D boxes, seg. \\
    RADIATE \cite{sheeny_radiate_2020} & 2020 & medium & \xmark & \xmark & \cmark & \xmark & \xmark  &  S  & CLO & \cmark & 2D boxes \\
    MulRan \cite{mulran_kim_2020} & 2020 &  medium  & \xmark & \xmark & \cmark & \cmark & \xmark &  S  &  CLO & \cmark  & \xmark  \\
    Zendar \cite{zendar} & 2020 & small  & \xmark & \xmark & \cmark & \cmark & \cmark & HD & CL & \cmark  & 2D boxes  \\
    CRUW \cite{rethinking_whang_2021} & 2021 & medium & \xmark & \xmark & \cmark & \xmark & \xmark  & LD & C & \cmark &  point location \\
    RadarScenes \cite{radarscenes_Schumann_2021} & 2021 &  large & \xmark & \xmark & \xmark & \cmark & \cmark & HD & CO  & \cmark & point-wise \\
    RADDet \cite{raddet_Zhang_2021} & 2021 &  small  & \xmark & \cmark & \cmark & \xmark & \cmark & LD & C  &  \cmark & 2D boxes \\
    \midrule
    \dataset~(ours) & 2022 & medium & \cmark & \cmark & \cmark & \cmark & \cmark & HD & CLO & \cmark  & 2D boxes, seg. \\
    \bottomrule
    \end{tabular}}
    \caption{\textbf{Publicly-available driving datasets with radar}. 
    The dataset is `small'  ($<$15k frames), `large'  ($>$130k frames) or `medium'  (in between). The radar is low-definition (`LD'), high-definition (`HD') or scanning (`S') and its data is released in different representations, amounting to different signal processing pipelines: analog-to-digital converter (`ADC') signal, range-azimuth-Doppler (`RAD') tensor, range-azimuth (`RA') view, range-Doppler (`RD') view, point cloud (`PC'). The presence of Doppler information depends on the radar sensor. Other sensor modalities are camera (`C'), LiDAR (`L') and odometry (`O').
    \dataset~is the only dataset providing each representation of a HD radar, combined with camera, LiDAR and odometry, while proposing detection and free-space segmentation tasks.}
    \label{tab:public_dataset}
\end{table*}

Automotive radars have been
in production since the late 90s. 
They are the preferred, most affordable sensors for adaptive cruise control, blind spot detection and automatic emergency braking functions. 
However, they have a poor angular resolution, which hinders their use in automated driving systems. 
Indeed, such systems need a high level of safety and robustness, usually reached through redundancy mechanisms. 
While sensing is improved by fusing several modalities, 
the overall combination works only if each sensor achieves sufficient and comparable performances. 
High-definition (HD) imaging radar has emerged to meet these requirements.
By using dense virtual antenna arrays, these new sensors achieve high angular resolution both in azimuth and elevation (horizontal and vertical angular positions, resp.) and produce denser point clouds.

With the rapid progress of deep learning and the availability of public driving datasets, \emph{e.g.}, \cite{Cityscapes, nuscenes_2020, kitti}, the perception ability of vision-based driving systems (detection of objects, structures, markings and signs, estimation of depth, forecasting of other road users' movements) has considerably improved. These advances quickly extended to depth sensors such as laser scanners (LiDAR), with the help of specific architectures to deal with 3D point clouds~\cite{pixor,pointpillars}. 

Quite surprisingly, the adoption of deep learning for radar processing in this context is much slower, compared to the other sensors. This might be explained by the complex nature of the data and the lack of public datasets. Indeed, recent key contributions in the field of radar-based vehicle's perception have appeared together with the release of datasets.
Interestingly, most of the recent works exploit the range-azimuth (RA) representation of the radar data (either in polar or Cartesian coordinates).
Similar to a bird's eye view (see Figure \ref{fig:teaser}d), this representation is easy to interpret and allows simple data augmentation with translations and rotations. However, one barely-mentioned drawback is that the generation of the RA radar maps incurs significant processing costs (tens of GOPS, see Section \ref{sec:complexity}), which compromises its viability on embedded hardware. While novel HD radars offer better resolution, they make this computational complexity issue even more acute.

Owing to the promising capabilities of HD radars, our work attacks this issue to improve their practicality. In particular, we propose:
(1) \method, an optimized  deep architecture that processes HD radar data at reduced cost, toward two different perception tasks, namely vehicle detection and free-space segmentation; 
(2) An empirical analysis comparing various radar signal representations in terms of performance, complexity and memory footprint;
(3) \dataset, the first raw HD radar dataset, including several other automotive-grade sensors, as described in Table \ref{tab:public_dataset}.

The paper is organized as follows: Sections \ref{sec:background} and \ref{sec:related_work} discuss radar background and related work; \method~and \dataset~are introduced in Sections \ref{sec:approach} and \ref{sec:dataset} resp.; Experiments are reported in Section \ref{sec:experiments}, and Section \ref{sec:conclusions} concludes.

\section{Radar background}
\label{sec:background}

Radars are usually composed of a set of transmitting and receiving antennas. 
The transmitters emit electromagnetic waves which are reflected back to the receivers by the objects in the environment.
Standard in the automotive industry \cite{brooker_understanding_2005, ghaleb_micro-doppler_2009}, a frequency-modulated continuous-wave (FMCW) radar emits a sequence of frequency-modulated signals called chirps. 
The frequency difference between the emission and reception is mostly due to the radial distance of the obstacle. 
This distance is thus extracted via a Fast Fourier Transform  (FFT) along the chirp sequence (range-FFT). 
A second FFT (Doppler-FFT) along the time axis extracts the phase difference, which captures the radial velocity of the reflector. 
The combination of these 2 FFTs provides a range-Doppler (RD) spectrum for each receiving antenna (Rx), stored for all Rx in an RD tensor.
The angle-of-arrival (AoA) can be estimated by using more than one Rx.
A phase difference in the received signal is observed due to the small distance between Rx antennas. 
A common practice is to apply a third FFT (angle-FFT) along the channel axis to estimate this AoA. 

Radar's capability to discriminate between two targets with same range and velocity but different angles is called its angular resolution. 
It is directly proportional to the antenna aperture, that is, the distance between the first and last antennas. 
The multiple inputs multiple outputs (MIMO) approach \cite{mimo} is commonly used to improve the angular resolution without increasing the physical aperture: Angular resolution increases by a factor of 2 for each added emitting antenna (Tx). 
Denoting $\Tx$ and $\Rx$ the number of its Tx and Rx channels respectively, a MIMO system builds a virtual array of $\Tx{\cdot}\Rx$ antennas. In order to prevent emitted signals from interfering, the transmitters emit the same signal at the same time, but with a slight phase shift $\Delta_{\phi}$ between two consecutive antennas. The downside of this approach is that the signature of each reflector appears $\Tx$ times in the RD spectrum, making the data interleaved.

To translate the AoA into an effective angle, one needs to calibrate the sensor. 
An alternative to the third FFT is to correlate in the complex domain the RD spectrum with a calibration matrix, to estimate the angles (azimuth and elevation). 
The complexity of this operation for a single point of the RD tensor is $\mathcal{O}(\Tx\Rx\BinA\BinE)$, where $\BinA$ and $\BinE$ are the numbers of discretization bins for azimuth and elevation angles respectively in the calibration matrix. 
For a 4D representation in range-azimuth-elevation-Doppler, this operation would need to be performed for each point of the RD tensor.\footnote{Considering a HD radar with $0.2^{\circ}$ of azimuth resolution over $180^{\circ}$ of horizontal field of view (FoV) and 11 elevations, it would require 498 GFLOPS to be computed.}

As a conclusion, for an embedded HD radar, traditional signal processing can not be applied as it is too resource greedy in terms of both computation requirements and memory footprint.
For driving assistance systems, there is therefore the challenge of increasing radar's angular resolution while keeping the processing costs under control.

\section{Related work}
\label{sec:related_work}

\noindent\textbf{Radar datasets.~} 
Traditional radars offer a good trade-off between cost and performance. While they provide accurate range and velocity, they suffer from a low azimuth resolution leading to ambiguity in separating close objects. 
Recent datasets include processed radar representations such as the entire range-azimuth-Doppler (RAD) tensor \cite{carrada_ouaknine_2021, raddet_Zhang_2021} or single views of this tensor --either range-azimuth (RA) \cite{barnes_oxford_2020, sheeny_radiate_2020, mulran_kim_2020, rethinking_whang_2021,zendar} or range-Doppler (RD) \cite{zendar}. 
These representations require large bandwidth to be transmitted as well as large memory storage. 
For this reason, datasets that include several modalities with numerous samples, such as nuScenes \cite{nuscenes_2020}, provide only radar point clouds, a lighter representation. However, it is a limited processed representation and it is biased to the signal processing pipeline. Several other datasets  use a $360^{\circ}$ scanning radar  \cite{barnes_oxford_2020, sheeny_radiate_2020, mulran_kim_2020}. However, its angular resolution is limited as with traditional radars and it does not provide Doppler information.

As discussed earlier, recent HD radars successfully reach an azimuth angular resolution below the degree using large arrays of virtual antennas. The Zendar dataset \cite{zendar} provides range-Doppler and range-azimuth views for such a radar. Both Astyx \cite{automotive_Meyer_2019} and RadarScenes \cite{radarscenes_Schumann_2021} datasets contain HD radar data processed as point clouds.

To the best of our knowledge, there is no open-source HD radar dataset that provides raw data together with camera and LiDAR in various driving environments, a gap that our dataset is filling.  
Table \ref{tab:public_dataset} summarizes the characteristics of the publicly-available driving datasets with radar.

\smallskip\noindent\textbf{Radar object detection.~} 
Low-definition (LD) radar has been used for many applications such as hand gesture recognition \cite{HandGesture}, object or person detection at gates \cite{Gate} and aerial monitoring \cite{Aerial}.
For automotive applications, single views of the RAD tensor are chosen as input of specific neural network architectures to detect objects' signatures in the considered view, either RA \cite{dong_probabilistic_2020, wang_rodnet_2021} or RD \cite{ng_range-doppler_2020}. Differently, \cite{zhang_object_2020} uses a radar view to localise objects in the camera image and 
\cite{brodeski_deep_2019} proposes a two-stage approach to estimate the azimuth of a detected object using only RD views.

Specific architectures have been designed to ingest aggregated views of the RAD tensor to detect objects in the RA view \cite{major_vehicle_2019,gao_ramp-cnn_2020}. 
The entire tensor has also been considered, either for object detection in both RA and RD views \cite{raddet_Zhang_2021} or for object localisation in the camera image \cite{palffy_cnn_2020}.

A radar point cloud contains less information than RAD views due to the pre-processing that has been applied. However, \cite{danzer_2d_2019, scheiner_seeing_2020} explore this representation for 2D object detection with LR radar and \cite{meyer_deep_2019} shows that HD radar point clouds can outperform LiDAR for this task.

None of these works mentions the pre-processing cost to generate the RAD tensor or the point cloud, which are taken as granted. 
In fact, a HD radar can not be used by previously-mentioned approaches as it would not fit on even the largest automotive embedded device. With \cite{gao_ramp-cnn_2020} for instance applied to HD radar, the input data at each timestamp would occupy 450MB and require $4.5{\cdot}10^{10}$ FLOPS\footnote{For comparison, ResNet50 on a $256$px image requires $4{\cdot}10^9$ FLOPS.} for only one elevation (out of 11).
To the best of our knowledge, there is no previous work on end-to-end object detection that is capable to scale with raw HD radar data.

\smallskip\noindent\textbf{Radar semantic segmentation.~}
Semantic segmentation on radar representation has been less explored due to the lack of annotated datasets. 
The RA view has been a research topic for multi-class \cite{kaul_rss-net_2020} and free-space \cite{nowruzi_polarnet_2021} segmentation. 
The entire RAD tensor is considered for multi-view segmentation in \cite{ouaknine_multi-view_2021}. 
Radar point cloud segmentation has also been explored to estimate bird-eye-view occupancy grids, either for LD \cite{lombacher_semantic_2017, sless_road_2019} or HD \cite{prophet_semantic_2019, prophet_semantic_2020, schumann_scene_2020} radars.

Once again, none of these methods can scale to raw HD radar data to perform free-space segmentation for instance.
Moreover, there is no previous work either on free-driving-space segmentation or semantic segmentation using only RD views of HD radar signals. 
In addition, there is no existing multi-task model that performs both radar object detection and semantic segmentation simultaneously.
Next, we detail our approach toward reducing both memory and complexity to perform vehicle detection and free-driving-space segmentation with raw HD radar signals.

\section{\method~architecture}
\label{sec:approach}

\begin{figure*}[t]
\renewcommand{\captionfont}{\small}
\centering
\includegraphics[width=0.95\linewidth]{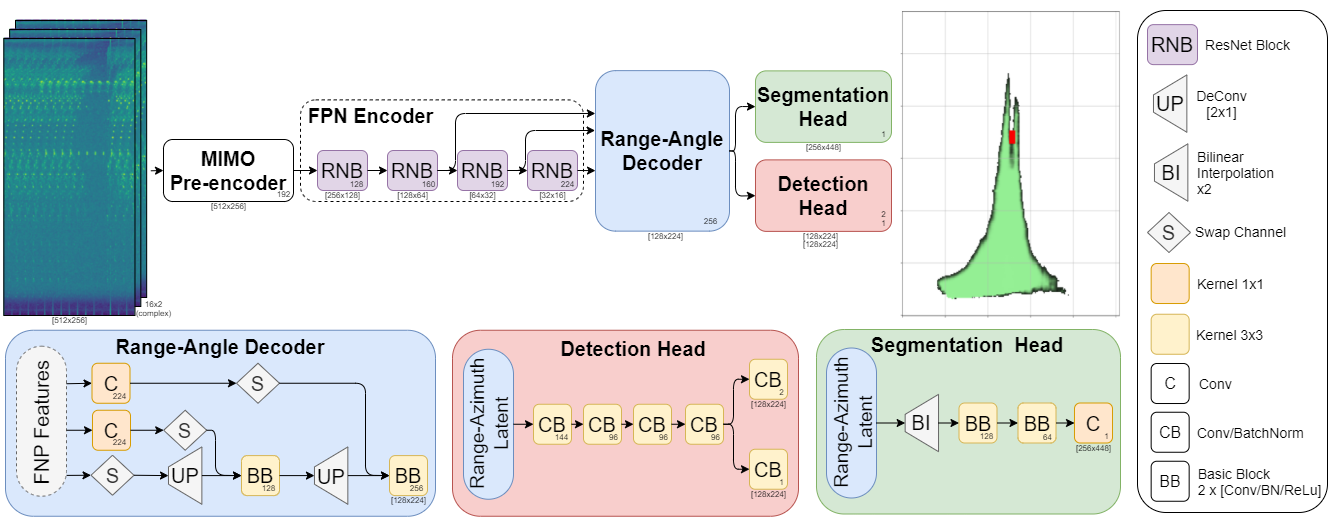}
  \caption{\textbf{Overview of \method}. \method~ is a lightweight multi-task architecture. It does not use any RA maps or RAD tensor which would require costly pre-processing. Instead, it leverages complex range-Doppler spectrums containing all the range, azimuth and elevation information. This data is de-interleaved and compressed by the MIMO pre-encoder. An FPN encoder extracts a pyramid of features which the range-angle decoder converts into a latent range-azimuth representation. Based on this representation, multi-task heads finally detect vehicles and predict the free driving space.
  } 
  \label{fig:arch}
\end{figure*}

Our approach is motivated by automotive constraints: Automotive-grade sensors must be used and only limited processing/memory resources are available on the embedded hardware. 
In this context, the RD spectrum is the only representation that is practical for HD radar. Based on it, we propose a multi-task architecture, compatible with above requirements, which is composed of five blocks (see Fig.\ref{fig:arch}):
\begin{itemize}[leftmargin=6pt,topsep=2pt,itemsep=0pt,parsep=0pt,partopsep=0pt]
  \item A pre-encoder reorganizing and compressing the RD tensor into a meaningful and compact representation;
  \item A shared Feature Pyramidal Network (FPN) encoder combining low-resolution semantic information with high-resolution details;
  \item A range-angle decoder building a range-azimuth latent representation from the feature pyramid;
  \item A detection head localizing vehicles in range-azimuth coordinates;
  \item A segmentation head predicting the free driving space.
\end{itemize}

\begin{figure}[t]
    \centering
    \includegraphics[width=1.1\columnwidth]{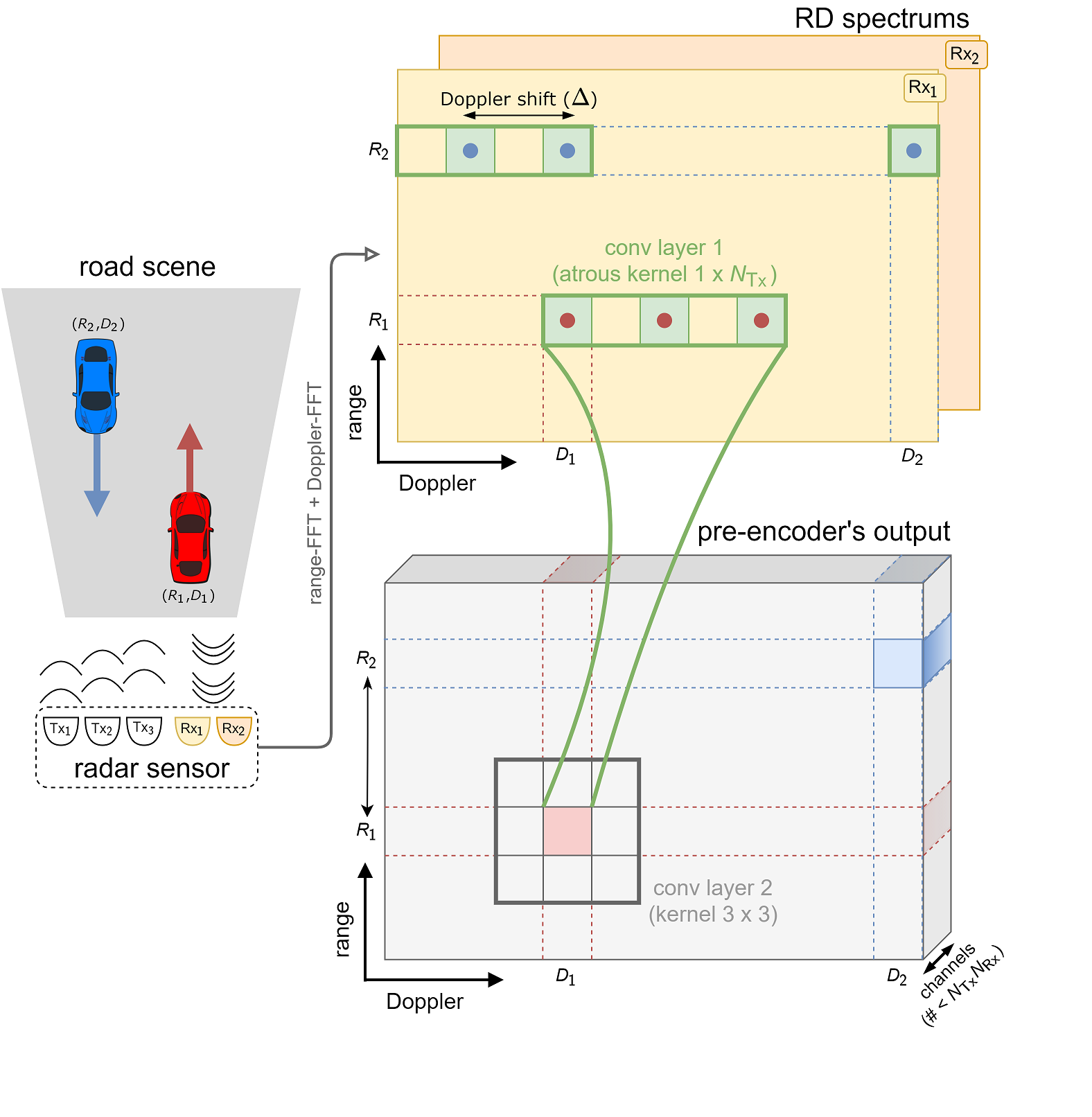}
    \vspace{-1.2cm}\caption{\textbf{Trainable MIMO pre-encoder.} Considering three transmitters ($\Tx\,{=}\,3$) and two receivers ($\Rx\,{=}\,2$), an object's signature is visible $\Tx$ times in the RD spectrum. The pre-encoder uses Atrous convolutions to organise and compress signatures in fewer than $\Tx{\cdot}\Rx$ output channels.
    }
    \label{mimo}
\end{figure}

\subsection{MIMO pre-encoder}
As explained in Section \ref{sec:background}, the MIMO configuration implies one complex RD spectrum per receiver.   
This results in a complex 3D tensor of dimension $(\BinR,\BinD,\Rx)$, where $\BinR$ and $\BinD$ are the numbers of discretization bins for range and Doppler respectively. 
It is important to understand how a given reflecting object, say a car in front, appears in this data. Denote $R$ the actual radial distance of this object to the radar and $D$ its relative radial velocity expressed in Doppler effect. For each receiver, its signature will be visible $\Tx$ times, one per transmitter. More specifically, it will be measured at range-Doppler positions $(R,(D+k\Delta)[D_\text{max}])_{k=1\cdots\Tx}$, where $\Delta$ is the Doppler shift (induced by the phase shift $\Delta_{\phi}$ in the transmitted signal) and $D_{\text{max}}$ is the largest Doppler that can be measured. The measured Doppler values are modulo this maximum.

This signal intricacy calls for a rearrangement of the RD tensor that will facilitate a subsequent exploitation of the MIMO information (to recover angles) while keeping data volume under control.
To this end, we propose a new trainable pre-encoder that performs such a compact reorganization of the input tensor  (Fig.\,\ref{mimo}). 
In order to handle at best its specific structure along the Doppler axis, we use first a suitably-defined Atrous convolution layer that gathers Tx and Rx information at the right positions.
The size of its kernel for one input channel is $1{\times}\Tx$, hence defined by the number of Tx antennas, and its dilation amounts to $\delta = \frac{\Delta \BinD}{D_{\text{max}}}$, the number of Doppler bins corresponding to Doppler shift $\Delta$. The number of input channels is the number $\Rx$ of Rx antennas.
A second convolution layer, with a $3{\times}3$ kernel, learns how to combine these channels and compresses the signal.
The two-layer pre-encoder is trained end-to-end with the rest of the proposed architecture. 

\subsection{FPN encoder}
Using a pyramidal structure to learn multi-scale features is a common practice in object detection \cite{FPN} and semantic segmentation \cite{PSPNet}.
Our FPN architecture uses 4 blocks composed of 3, 6, 6 and 3 residual layers \cite{residual} respectively.
The feature maps of these residual blocks form the feature pyramid. This classic encoder has been optimized considering the nature of the data while controlling its complexity. 
The channel dimensions are in fact chosen to encode at best the azimuth angle over the entire distance range (\textit{i.e.}, high resolution and narrow field of view at far range, low resolution and wider field of view at near range). To prevent losing the signature of small objects (typically few pixels in the RD spectrum), the FPN encoder performs a $2{\times}2$ down-sampling per block, leading to a total reduction of the tensor size by a factor of 16 in height and width.
For similar reasons and to avoid overlaps between adjacent Tx's, it uses $3{\times}3$ convolution kernels.

\subsection{Range-angle decoder}
The range-angle decoder aims to expand the input feature maps to higher resolution representations.
This upscaling is usually achieved through multiple deconvolution layers whose output is combined with previous feature maps to preserve spatial details. 
In our case, the representation is unusual due to the physical nature of the axes: The dimensions of the input tensor correspond respectively to range, Doppler and azimuth angle, whereas the feature maps that will be sent to subsequent task heads should correspond to a range-azimuth representation. 
Consequently, we swap the Doppler and azimuth axes to match the final axis ordering and then upscale the feature maps. 
However, the range axis has a lower size compared to the azimuth one, since it was decimated by a factor of 2 after each residual block, while the azimuth axis (formerly the channel axis) was increasing. 
Prior to these operations, we apply a $1{\times}1$ convolution to the feature maps from the encoder to the decoder. It adjusts the dimension of the azimuth channel to its final size, right before swapping the axes. The deconvolution layers upscale only the range axis, producing feature maps that are concatenated with those from the previous pyramid level. 
A final block of two Conv-BatchNorm-ReLU layers is applied, generating the final range-azimuth latent representation.

\subsection{Multi-task learning}
\label{sec:multi-task}

\noindent\textbf{Detection task.~}
The detection head is inspired from Pixor \cite{pixor}, an efficient and scalable single-stage model. 
It takes the RA latent representation as input and processes it using a first common sequence of four Conv-BatchNorm layers with 144, 96, 96 and 96 filters respectively.
The branch is then divided in a classification and a regression pathways.
The classification part is a convolution layer with sigmoid activation that predicts a probability map.
This output corresponds to binary classification of each ``pixel'' as occupied or not by a vehicle. 
In order to reduce computational complexity, it predicts a coarse RA map, where each cell has a resolution of $0.8$m in range and $0.8^{\circ}$ in azimuth (\emph{i.e.}, $\sfrac{1}{4}$ and $\sfrac{1}{8}$ of native resolutions resp. in range and azimuth). This cell size is enough to dissociate two close objects. Then, the regression part finely predicts the range and azimuth values corresponding to the detected object. To do so, a unique $3{\times}3$ convolution layer outputs two feature maps corresponding to the final range and azimuth values. 

This two-fold detection head is trained with a multi-task loss composed of a focal loss applied to all the locations for the classification and of a ``smooth L1'' loss for the regression applied only on positive detections (see \cite{pixor} for details of these losses).
Let $\vx$ be a training example, $\vy_{\text{clas}} \in \{ 0, 1\}^{\sfrac{\BinR}{4}{\times}\sfrac{\BinA}{8}}$ its classification ground truth and $\vy_{\text{reg}} \in \mathbb{R}^{2{\times}\sfrac{\BinR}{4}{\times}\sfrac{\BinA}{8}}$ the associated regression ground truth. 
The detection head of \method~predicts a detection map $\hat{\vy}_{\text{clas}} \in [0, 1]^{\sfrac{\BinR}{4}{\times}\sfrac{\BinA}{8}}$ and associated regression map $\hat{\vy}_{\text{reg}}\in \mathbb{R}^{2{\times}\sfrac{\BinR}{4} \times \sfrac{\BinA}{8}}$. Its training loss reads:
\begin{equation}\begin{split}
    \mathcal{L}_{\text{det}}(\vx,\vy_{\text{clas}},\vy_{\text{reg}}) =
    &~\text{focal}(\vy_{\text{clas}}, \hat{\vy}_{\text{clas}}) + \\ 
    &~\beta\,\text{smooth-L1}(\vy_{\text{reg}} - \hat{\vy}_{\text{reg}}),
\end{split}
\end{equation}
where $\beta>0$ is a balancing hyper-parameter. 

\smallskip\noindent\textbf{Segmentation task.~}
The free-driving-space segmentation task is formulated as a pixel-level binary classification. The segmentation mask has a resolution of $0.4$m in range and $0.2^{\circ}$ in azimuth. It corresponds to half of the native range and azimuth resolutions  while considering only half of the entire azimuth FoV (within $[-45^\circ, 45^\circ]$). The RA latent representation is processed by two consecutive groups of two Conv-BatchNorm-ReLu blocks, producing respectively 128 and 64 feature maps. A final $1{\times}1$ convolution outputs a 2D feature map followed by a sigmoid activation to estimate the probability of each location to be drivable. 
Let $\vx$ be a training example, $\vy_{\text{seg}} \in \{ 0, 1\}^{\sfrac{\BinR}{2}{\times}\sfrac{\BinA}{4}}$ its one-hot ground truth and $\hat{\vy}_{\text{seg}} \in \left[ 0, 1\right]^{\sfrac{\BinR}{2}{\times}\sfrac{\BinA}{4}}$ the predicted soft detection map. 
The segmentation task is learnt using a Binary Cross Entropy loss:
\begin{equation}
    \mathcal{L}_{\text{free}}(\vx,\vy_{\text{seg}}) = \sum_{(r, a) \in \Omega} \text{BCE}(\vy_{\text{seg}}(r,a),\hat{\vy}_{\text{seg}}(r,a)),
\end{equation}
where $\Omega = \llbracket 1,\frac{\BinR}{2}\rrbracket \times \llbracket 1,\frac{\BinA}{4}\rrbracket$.

\smallskip\noindent\textbf{End-to-end multi-task training.~}
The whole \method~model is trained by minimizing a combination of the previous detection and segmentation losses:
\begin{equation}
    \mathcal{L}_{\text{MTL}} = \sum_{\vx} \mathcal{L}_{\text{det}}(\vx,\vy_{\text{clas}},\vy_{\text{reg}}) + \lambda \mathcal{L}_{\text{free}}(\vx,\vy_{\text{seg}}),
\end{equation}
w.r.t. the parameters of the MIMO pre-encoder, of the FPN encoder, of the RA decoder and of the two heads; $\lambda$ is a positive hyper-parameter that balances the two tasks.

\section{\dataset~dataset}
\label{sec:dataset}

As depicted in Table\,\ref{tab:public_dataset}, publicly-available  datasets do not provide raw radar signal, neither for LD radar nor for HD radar. Therefore, we built \dataset, a new dataset to allow research on automotive HD radar.
As \dataset~includes 3 sensor modalities --camera, radar and laser scanner--, it should also permit one to investigate the fusion of HD radar with other more classic sensors. The specifications of the used sensor suite are detailed in Appendix \ref{sec:append_dataset}.   
Except for the camera, all sensors are automotive-grade qualified. 
On top of that, the GPS position and full CAN bus of the vehicle (including odometry) are also provided. 
Sensor signals were recorded simultaneously in a raw format, without any signal pre-processing. 
In the case of the HD radar, the raw signal is the ADC. From this ADC data, all conventional radar representations can be generated: range-azimuth-Doppler tensor, range-azimuth and range-Doppler views or point cloud.

\dataset~contains 91 sequences of about 1-4 minutes, for a total of 2 hours. This amounts to approximately 25k synchronized frames in total, out of which 8,252 are labelled with 9,550 vehicles (see details in Appendix \ref{sec:append_dataset}). 
Vehicles’ annotation is composed of 2D boxes in the image plane along with the real-world distance to the sensor and the Doppler value (relative radial speed). 
The annotation of the radar signal is hard to achieve as the RD spectrum representation is not meaningful for the human eye. 

Vehicle detection labels were first generated automatically using supervision from the camera and laser scanner. 
A RetinaNet model \cite{focal_loss} was used to extract object proposals from the camera. Then, these proposals were validated when both radar and LiDAR agree on the object position from their respective point cloud. Finally, manual verification was conducted to reject or validate the labels. 
The free-space annotation was done fully automatically on the camera images. A DeepLabV3+ \cite{chen_encoder-decoder_2018}, pre-trained on Cityscape, has been fine-tuned with 2 classes (\textit{free space} and \textit{occupied}) on a small manually-annotated part of our dataset.
This model segmented each video frame and the obtained segmentation mask was projected from the camera's coordinate system to the radar's one thanks to known calibration. 
Finally, already available vehicle bounding boxes were subtracted from the free-space mask. 
The quality of the segmentation mask is limited due to the automatic method we employed and to the projection inaccuracy from camera to real world.

\section{Experiments}
\label{sec:experiments}

\begin{table*}[!t]
    \scriptsize
    \centering
    \resizebox{\linewidth}{!}{
\begin{tabular}{l | c|  c c c c| c c c c |c c c c}
\toprule
      \multirow{2}{*}{Model} & 
      Radar
      &\multicolumn{4}{c|}{Overall} & \multicolumn{4}{c|}{Easy} & \multicolumn{4}{c}{Hard}\\

 & Input & AP($\%$) $\uparrow$ & AR($\%$) $\uparrow$ & R(m) $\downarrow$ &  A($^{\circ}$) $\downarrow$ & AP($\%$) $\uparrow$ & AR($\%$) $\uparrow$ & R(m) $\downarrow$ &  A($^{\circ}$) $\downarrow$ & AP($\%$) $\uparrow$ & AR($\%$) $\uparrow$ & R(m) $\downarrow$ &  A($^{\circ}$) $\downarrow$\\
\midrule

    Pixor \cite{pixor} & PC & 96.46 & 32.32 & 0.17 & 0.25 & $\textbf{99.02}$ & 28.83 & 0.15 & 0.19 & 93.28 & 38.69 & 0.19 & 0.33\\
    Pixor \cite{pixor} & RA & 96.56 & 81.68 & $\textbf{0.10}$ & 0.20 &
              96.86 & 88.02 & $\textbf{0.09}$ & 0.16 &
              $\textbf{95.88}$ & $\textbf{70.10}$ & $\textbf{0.12}$ & 0.27\\
    FFT-RadNet (ours) & RD & $\textbf{96.84}$ & $\textbf{82.18}$ & 0.11 & $\textbf{0.17}$	&
                            98.49 & $\textbf{91.69}$ & 0.10 & $\textbf{0.13}$ & 
                            92.93 & 64.82 & 0.13 & $\textbf{0.26}$\\ \bottomrule
                            
    \end{tabular}
    }
    \caption{\label{tab:Perf_objects}\textbf{Object detection performances on \dataset~Test split}.
    Comparison between Pixor trained with point cloud (`PC') or range-azimuth (`RA') representations, and the proposed \method~requiring only range-Doppler (`RD') as input. Our method obtains similar or better overall performances than baselines in both average precision (`AP') and average recall (`AR') for a 50\% IoU threshold. It also reaches similar or better range (`R') and angle (`A') accuracy, showing that it successfully learns a signal processing pipeline that estimates the AoA with significantly fewer operations, as detailed in Table \ref{tab:complexity}.
    }
\end{table*}

\subsection{Training details}

The proposed architecture has been trained on the \dataset~ dataset using exclusively the RD spectrums as input. The RD spectrum being composed of complex numbers, we stack its real and imaginary parts along the channel axis before passing it to the MIMO pre-encoder.
The dataset has been split into Training, Validation and Test sets (approx. 70\%, 15\% and 15\% of the dataset, respectively) in such a way that frames from a same sequence can not appear in different sets.
We manually split the Test dataset in `hard' and `easy' cases. Hard cases are mostly situations where the radar signal is perturbed, \emph{e.g.}, by interference with other radars, important side-lobes effects or significant reflections on metallic surfaces. 

The \method~architecture is trained using the multi-task loss detailed in Section \ref{sec:multi-task} with the following hyper-parameters set-up empirically: $\lambda=100$, $\beta=100$ and $\gamma=2$.
The training process uses the Adam optimizer \cite{kingma_adam_2015} during 100 epochs, with an initial learning rate of  $10^{-4}$ and a decay of 0.9 every 10 epochs.

\begin{figure*}[t]
\renewcommand{\captionfont}{\small}
\centering
\scriptsize
  \resizebox{0.98\textwidth}{!}{\begin{tabular}{c c c c  c c c}
    &\multicolumn{3}{c}{Easy} & \multicolumn{3}{c}{Hard}\\
    \cmidrule(r){2-4}\cmidrule(l){5-7}
    \raisebox{2.0\normalbaselineskip}[0pt][0pt]{\rotatebox[origin=c]{90}{Camera}} &
    \includegraphics[width=2cm]{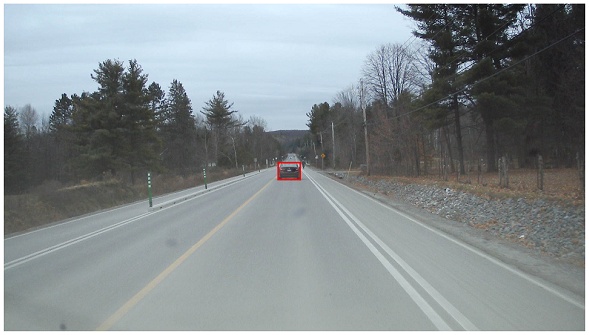} & 
    \includegraphics[width=2cm]{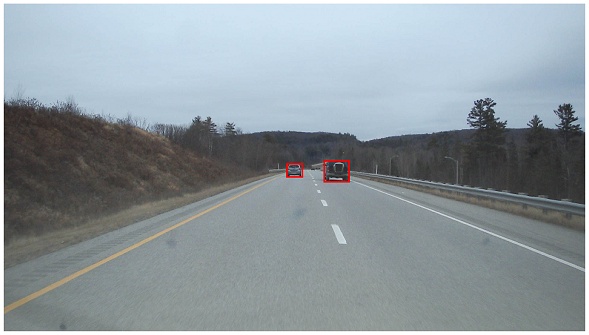} & 
    \includegraphics[width=2cm]{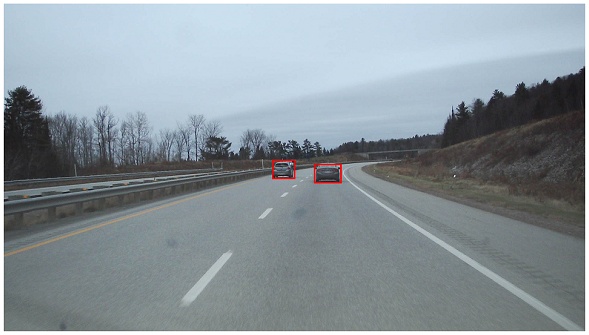} & 
    \includegraphics[width=2cm]{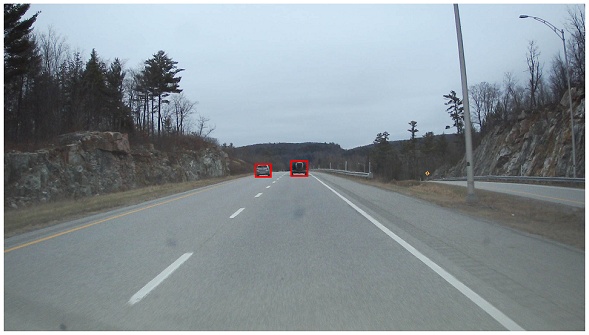} & 
    \includegraphics[width=2cm]{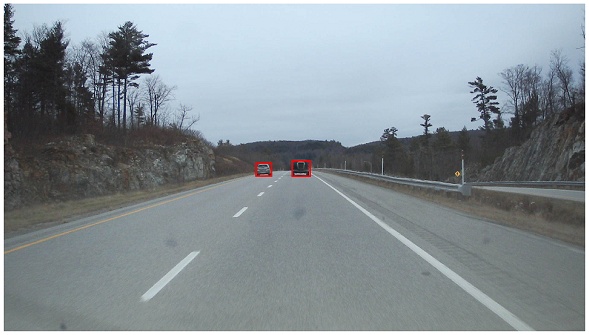} & 
    \includegraphics[width=2cm]{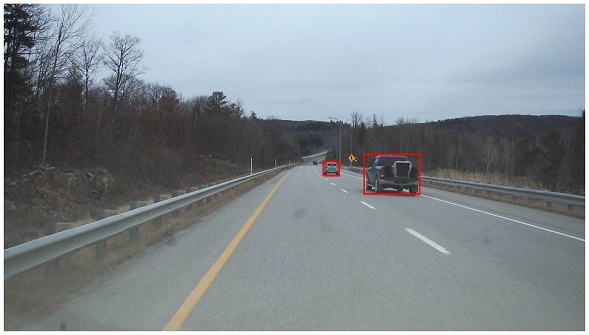}\\
    
    \raisebox{4\normalbaselineskip}[0pt][0pt]{\rotatebox[origin=c]{90}{Radar Input}}  &
    \includegraphics[width=2cm,height=2.5cm]{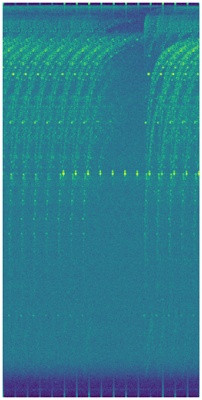} & 
    \includegraphics[width=2cm,height=2.5cm]{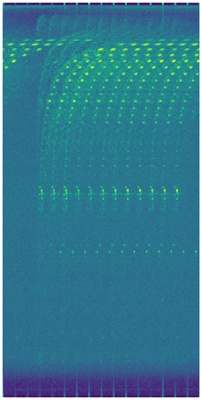} & 
    \includegraphics[width=2cm,height=2.5cm]{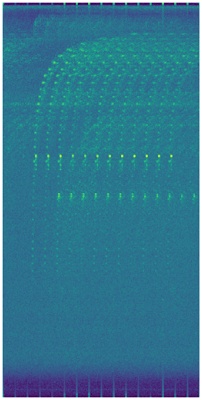} & 
    \includegraphics[width=2cm,height=2.5cm]{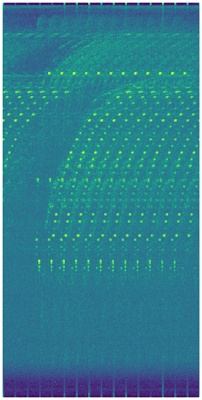} & 
    \includegraphics[width=2cm,height=2.5cm]{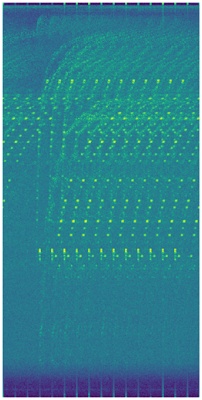} & 
    \includegraphics[width=2cm,height=2.5cm]{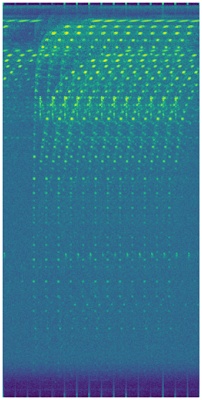}\\
    
    \raisebox{3.5\normalbaselineskip}[0pt][0pt]{\rotatebox[origin=c]{90}{Ground Truth}}
    &
    \includegraphics[width=2cm]{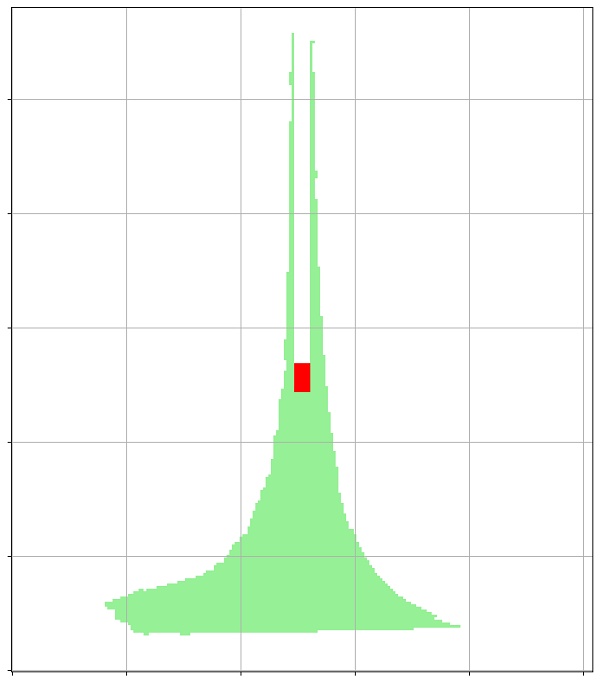} & 
    \includegraphics[width=2cm]{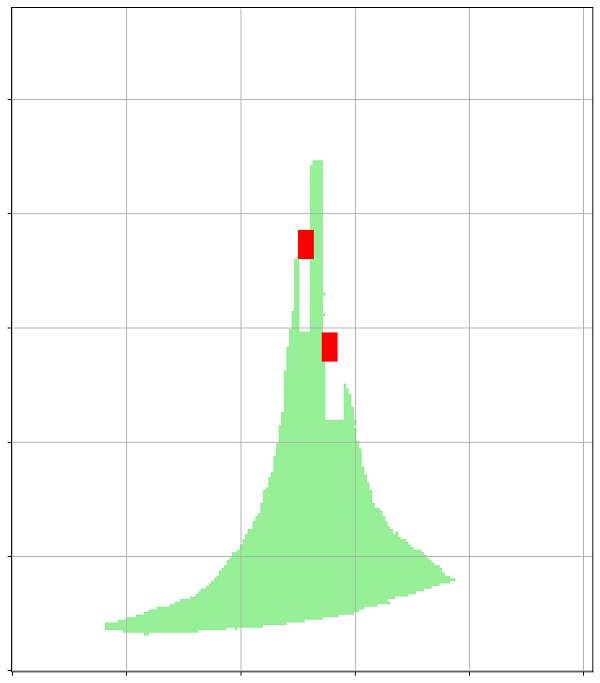} & 
    \includegraphics[width=2cm]{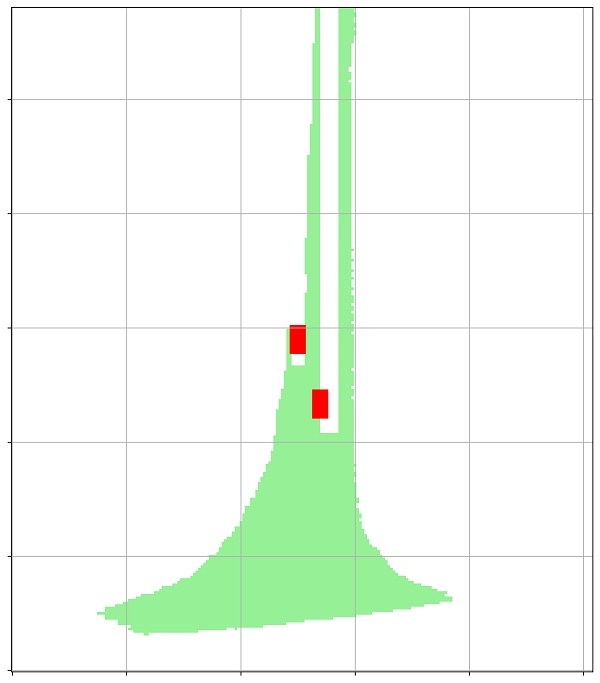} & 
    \includegraphics[width=2cm]{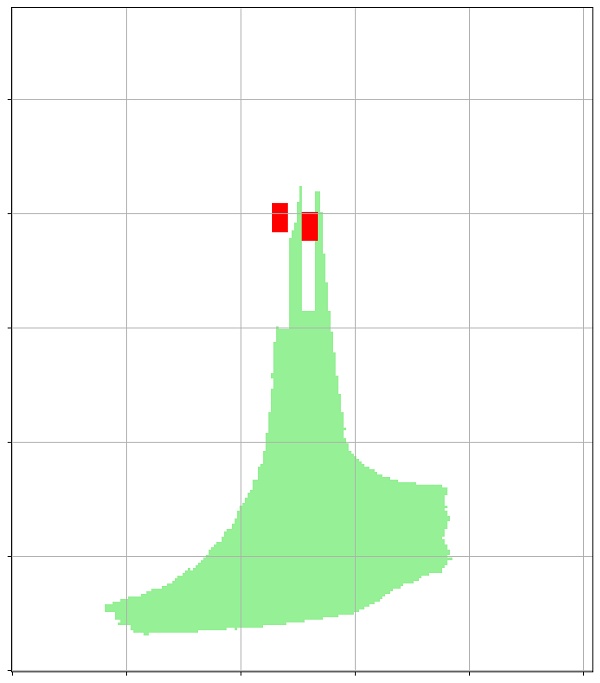} & 
    \includegraphics[width=2cm]{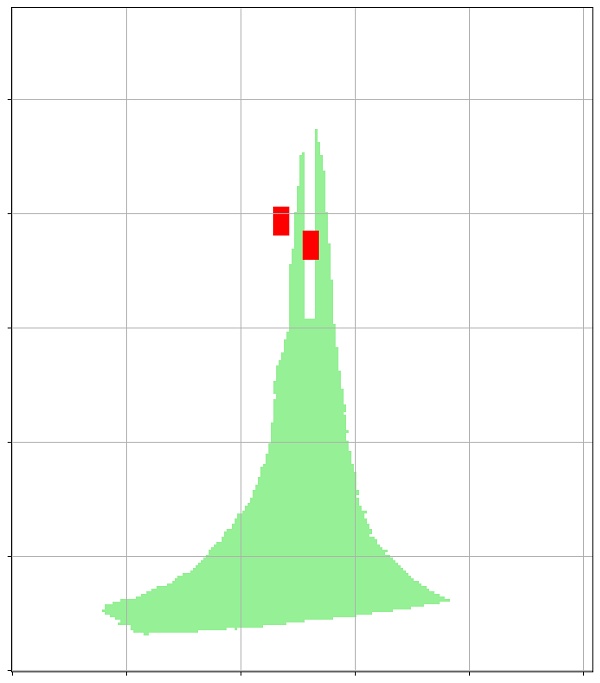} & 
    \includegraphics[width=2cm]{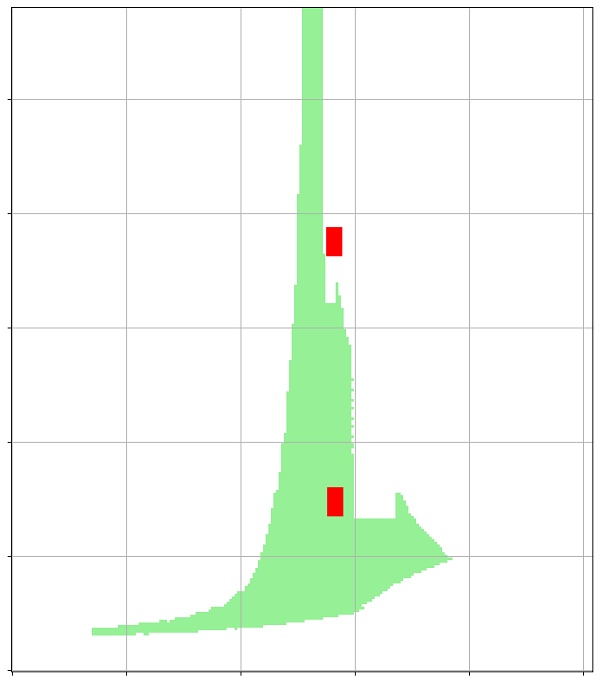}\\
    
    \raisebox{3.5\normalbaselineskip}[0pt][0pt]{\rotatebox[origin=c]{90}{Prediction}}
    &
    \includegraphics[width=2cm]{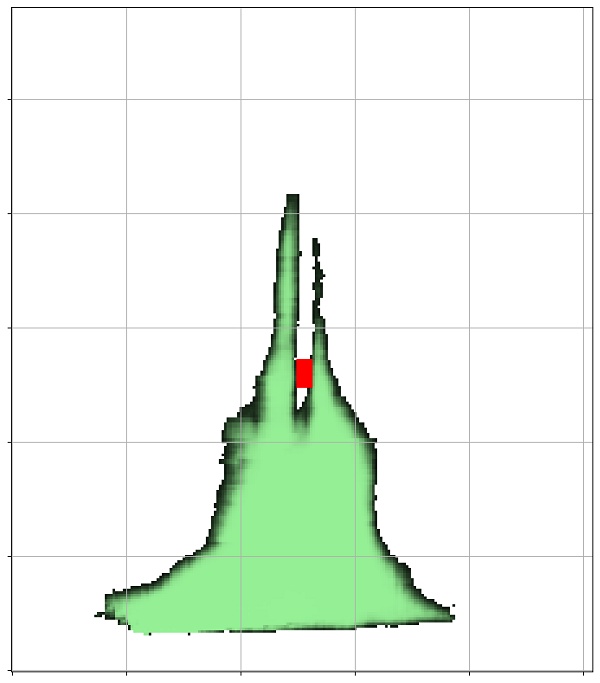} & 
    \includegraphics[width=2cm]{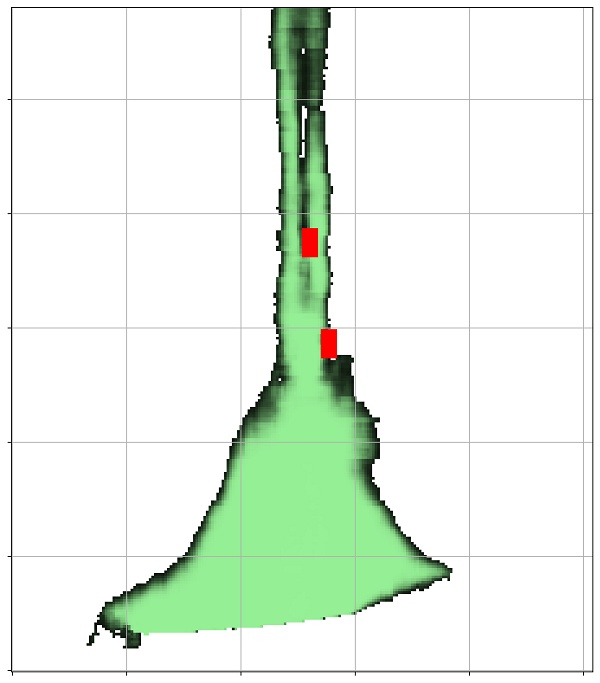} & 
    \includegraphics[width=2cm]{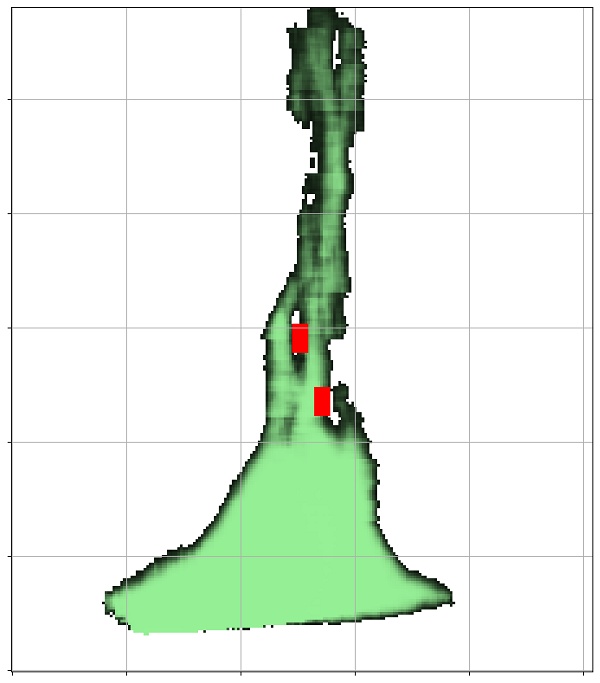} & 
    \includegraphics[width=2cm]{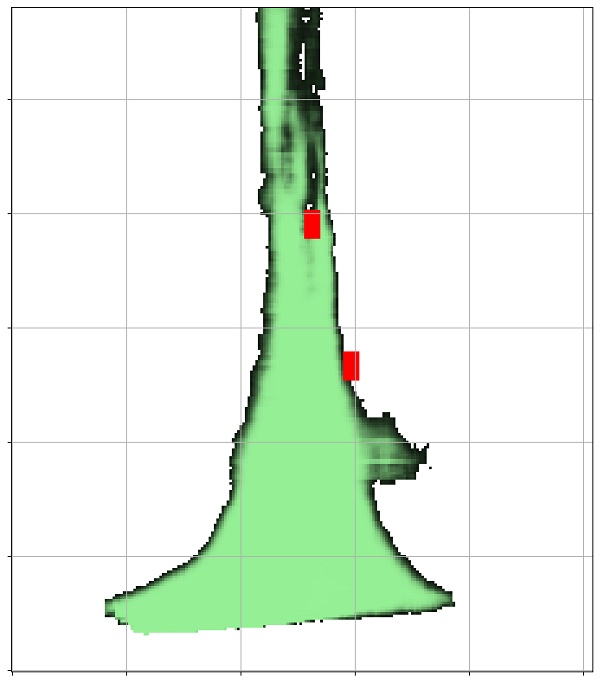} & 
    \includegraphics[width=2cm]{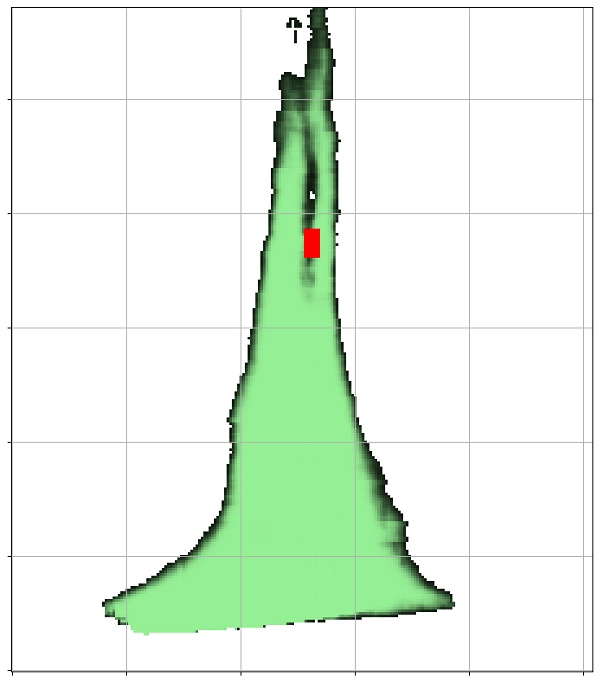} & 
    \includegraphics[width=2cm]{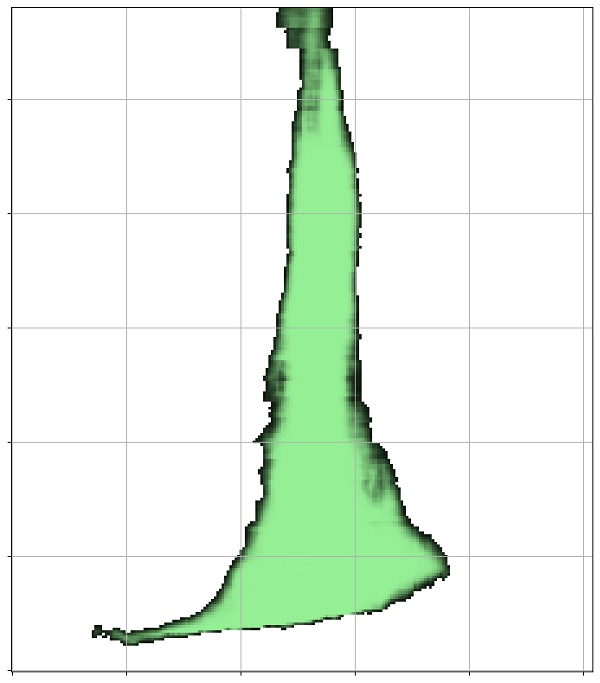}\\
    
  \end{tabular}}
  \caption{
  \textbf{Qualitative results for object detection and free-space segmentation on Easy and Hard samples.} Camera views (1st row) are displayed for visual reference only; RD spectrums (2nd row) are the only inputs to the model; Ground truths (3rd row) and predictions (4th row) are shown for both tasks. Note that there could be a projection error of the free driving space from camera to real world due to vehicle pitch variations.}
\label{fig:qualitative}
\end{figure*}

\subsection{Baselines}

The proposed architecture has been compared to recent contributions in the radar community.
Most of the competing methods presented in Section \ref{sec:related_work} have been designed for LD radar and can not scale with HD radar data due to memory limitation.
Instead, baselines with similar complexity have been selected regarding their input representation (range-azimuth or point cloud) for a fair comparison.
Input representations (RD, RA or point cloud) are generated for the entire Training, Validation and Test sets using a conventional signal-processing pipeline.

\smallskip\noindent \textbf{Object detection with point cloud.} 
The Pixor \cite{pixor} method has been used to detect vehicles after voxelization of the radar point cloud into a 3D volume of 
$[0\,\text{m}, 103\,\text{m}] {\times} [−40 \,\text{m}, 40 \,\text{m}] {\times}[-2.5 \,\text{m}, 2.0 \,\text{m}]$ around the radar (longitudinal, lateral and vertical ranges), sampled at $0.1$m in each direction. 
The size for this input 3D grid is thus $1030\!\times\! 800\!\times\! 45$. Pixor is a lightweight architecture intended to be real-time. However, its input representation generates 96MB of data, which becomes a challenge for embedded devices.

\smallskip\noindent \textbf{Object detection with RA tensor.}
As detailed in Section \ref{sec:related_work}, several methods \cite{major_vehicle_2019, gao_ramp-cnn_2020} used views of the RAD tensor as input. However, the memory usage would be too extensive for HD radar data. As \cite{major_vehicle_2019} showed that using only the RA view leads to better performance for object detection, we compared our method to a Pixor architecture without the voxelization module. It takes as input the RA representation in \dataset, of size $512{\times}896$ with range values in $[0 \text{m}, 103 \text{m}]$ and azimuth in $[-90^\circ, 90^\circ]$.

\smallskip\noindent\textbf{Free-space segmentation.} We selected PolarNet \cite{nowruzi_polarnet_2021} to evaluate against our approach. It is a lightweight architecture designed to process RA maps and predict free space. We re-implemented it to the best of our comprehension.

\subsection{Evaluation metric}
For object detection, the Average Precision (AP) and Average Recall (AR) are used considering an Intersection-Over-Union (IoU) threshold of 50\%. 
For semantic segmentation, the mean IoU (mIoU) metric is used on a binary classification task (\textit{free} or \textit{occupied}). The metric is computed on a reduced $[0 \text{m},50 \text{m}]$ range as the boundaries of the road surface are hardly visible beyond this distance.

\subsection{Performance analysis}

\noindent \textbf{Object detection.} 
Performances for object detection are reported in Table \ref{tab:Perf_objects}. We observe that \method~using range-Doppler as input outperforms the Pixor baseline using PC as input (Pixor-PC) and reaches slightly better performances than the costly Pixor-RA baseline. 
The position accuracy, both in range and azimuth angle, is similar, and even better in angle, compared to Pixor-RA. These results show that our approach successfully learns the azimuth angle from the data. From a manufacturing viewpoint, note that this opens cost saving opportunities as the end-of-line calibration of the sensor is no longer required in the proposed framework. 
In the Easy Test set, \method~delivers +1.6\% AP and +3.6\% AR compared to Pixor-RA. However, on the Hard test set, Pixor-RA performs the best.  
The RA approach does not struggle that much with the hard samples because the data is pre-processed by a signal processing pipeline that already solves some of these cases. 
In contrast, the performance with point-cloud input is much lower than all others. Indeed, the recall is low due to the limited number of points at far range. 

\smallskip\noindent \textbf{Free-driving-space segmentation.} The performance for the free-driving-space segmentation is provided in Table \ref{tab:Perf_freespace}. We observe that \method~significantly outperforms PolarNet by 13.4\% IoU on average. This is partly explained by the lack of elevation information in the RA map, an information that is present in the RD spectrums. 

\subsection{Complexity analysis}
\label{sec:complexity}
\method~has been designed first to get rid of the signal processing chains that transform the ADC data into either a sparse point cloud or denser representations (RA or RAD), without compromising the richness of the signal.
Because the input data remains quite large, we designed a compact model to bound the complexity in terms of number of operations, as a trade-off between performance and range/angle accuracy. Moreover, the pre-encoder layer compresses significantly the input data. An ablation study has been performed to define the best trade-off between the size of the feature maps and the model's performance (details in Appendix \ref{sec:append_ablation}).

\begin{table}[t]
    \centering
    \resizebox{0.90\width}{!}{
\begin{tabular}{l  c  c  c  c}
\toprule
\multirow{2}{*}{Model} & \multirow{2}{*}{Radar input} & \multicolumn{3}{c}{mIoU (\%) $\uparrow$}\\ \cmidrule(l){3-5}
&  & Overall & Easy & Hard\\
\midrule
    PolarNet \cite{nowruzi_polarnet_2021} & RA & 60.6 & 61.9 & 57.4 \\
    FFT-RadNet & RD & \textbf{74.0} & \textbf{74.6} & \textbf{72.3} \\
    \bottomrule
    \end{tabular}
    }
    \caption{\label{tab:Perf_freespace}\textbf{Free-driving-space segmentation performances}. \method~successfully approximates the angle information in the radar data while reaching better performance than PolarNet. Note that this performance is achieved by \method~while simultaneously performing object detection, as our model is multi-task.}
\end{table}

\begin{table}[t]
    \centering
\resizebox{\linewidth}{!}{%
\begin{tabular}{l  c c c c}
\toprule
    Method & \multirow{2}{*}{\shortstack{Input size \\ (MB)  $\downarrow$}} & \# \multirow{2}{*}{\shortstack{Params. \\ ($10^{6}$) $\downarrow$}} & \multicolumn{2}{c}{Complexity (GFLOPS) $\downarrow$}\\ \cmidrule{4-5}
    & & & AoA proc. & Model \\
\midrule
    PCL Pixor  & 98.30 & 6.93 & 8 & 741\\
    RA Pixor   & $~~\textbf{1.75}$ & 6.92 & 45* & 761\\
    FFT-RadNet  & 16.00 & $\textbf{3.79}$ & $\textbf{0}$ & $\textbf{584}$\\ \bottomrule

    \end{tabular}%
    }
    \caption{\textbf{Complexity analysis}. The proposed method reaches the best trade-off between the size of the input, the number of parameters of the model and the computational complexity. Note that the AoA processing of the RA Pixor method (*) considers only a single elevation, otherwise it is up to 496 GFLOPS for the whole set of $\BinE\! =\! 11$ elevations.}
    \label{tab:complexity}
\end{table}

As shown in Tab.\,\ref{tab:complexity}, \method~is the only method not requiring the AoA estimation. As explained in Sec.\,\ref{mimo}, the pre-encoder layer compresses the MIMO signal containing all the information to recover both the azimuth and elevation angles. The AoA for the point-cloud approach generates 3D coordinates for a sparse cloud of around 1000 points on average, leading to 8 GFLOPS-worth of computing, prior to applying Pixor for object detection. 
To produce the RA or RAD tensor, AoA runs for each single bin of the RD map, but only considering one elevation. Such a model is thus unable to estimate the elevation of objects such as bridges or lost cargo (low object). For one elevation, the complexity is about 45 GFLOPS, but would increase up to 495 GLPOPS for all the 11 elevations. We have demonstrated that \method~can cut these processing costs without compromising the quality of the estimation.

\section{Conclusion}
\label{sec:conclusions}

We introduced \method, a novel trainable architecture to process and analyse HD radar signals. We demonstrated that it effectively alleviates the need for costly pre-processing to estimate RA or RAD representations. Instead, it detects and estimates objects position while segmenting free driving space from the RD spectrum directly. \method~slightly outperforms RA-based approaches 
while reducing processing requirements. The experiments are conducted on
\dataset, a new dataset which is part of the work and contains sequences of automotive-grade sensor signals (HD radar, camera and laser scanner).  Synchronized sensor data are available in a raw format so that various representations can be evaluated and further research can be conducted, possibly with fusion-based approaches.

{\small
\bibliographystyle{ieee_fullname}
\bibliography{CVPR_bib}

\begin{thebibliography}{10}\itemsep=-1pt

\bibitem{barnes_oxford_2020}
Dan Barnes, Matthew Gadd, Paul Murcutt, Paul Newman, and Ingmar Posner.
\newblock The {Oxford} {Radar} {RobotCar} {dataset}: {a} {radar} {extension} to
  the {Oxford} {RobotCar} {dataset}.
\newblock In {\em {ICRA}}, 2020.

\bibitem{brodeski_deep_2019}
Daniel Brodeski, Igal Bilik, and Raja Giryes.
\newblock Deep {radar} {detector}.
\newblock In {\em RadarConf}, 2019.

\bibitem{brooker_understanding_2005}
Graham~M Brooker.
\newblock Understanding millimetre wave fmcw radars.
\newblock In {\em ICST}, 2005.

\bibitem{nuscenes_2020}
Holger Caesar, Varun Bankiti, Alex~H. Lang, Sourabh Vora, Venice~Erin Liong,
  Qiang Xu, Anush Krishnan, Yu Pan, Giancarlo Baldan, and Oscar Beijbom.
\newblock {nuScenes}: {a} {multimodal} {dataset} for {autonomous} {driving}.
\newblock In {\em {CVPR}}, 2020.

\bibitem{chen_encoder-decoder_2018}
Liang-Chieh Chen, Yukun Zhu, George Papandreou, Florian Schroff, and Hartwig
  Adam.
\newblock Encoder-{decoder} with {atrous} {separable} {convolution} for
  {semantic} {image} {segmentation}.
\newblock In {\em ECCV}, 2018.

\bibitem{Cityscapes}
Marius Cordts, Mohamed Omran, Sebastian Ramos, Timo Rehfeld, Markus Enzweiler,
  Rodrigo Benenson, Uwe Franke, Stefan Roth, and Bernt Schiele.
\newblock The cityscapes dataset for semantic urban scene understanding.
\newblock In {\em CVPR}, 2016.

\bibitem{danzer_2d_2019}
Andreas Danzer, Thomas Griebel, Martin Bach, and Klaus Dietmayer.
\newblock {2D} {car} {detection} in {radar} {data} with {PointNets}.
\newblock In {\em ITSC}, 2019.

\bibitem{dong_probabilistic_2020}
Xu Dong, Pengluo Wang, Pengyue Zhang, and Langechuan Liu.
\newblock Probabilistic {oriented} {object} {detection} in {automotive}
  {radar}.
\newblock In {\em CVPR Worshops}, 2020.

\bibitem{mimo}
B~J Donnet and I~D Longstaff.
\newblock Mimo radar, techniques and opportunities.
\newblock In {\em EuRAD}, 2006.

\bibitem{HandGesture}
S. Franceschini, M. Ambrosanio, S. Vitale, F. Baselice, A. Gifuni, G. Grassini,
  and V. Pascazio.
\newblock Hand gesture recognition via radar sensors and convolutional neural
  networks.
\newblock In {\em RadarConf}, 2020.

\bibitem{gao_ramp-cnn_2020}
Xiangyu Gao, Guanbin Xing, Sumit Roy, and Hui Liu.
\newblock {RAMP}-{CNN}: {a} {novel} {neural} {network} for {enhanced}
  {automotive} {radar} {object} {recognition}.
\newblock In {\em Sensors}, 2020.

\bibitem{kitti}
Andreas Geiger, Philip Lenz, Christoph Stiller, and Raquel Urtasun.
\newblock Vision meets robotics: The kitti dataset.
\newblock {\em IJRR}, 2013.

\bibitem{ghaleb_micro-doppler_2009}
Antoine Ghaleb.
\newblock {\em Micro-{Doppler} analysis of non-stationary moving targets in
  radar imaging}.
\newblock PhD thesis, Telecom Paris, 2009.

\bibitem{residual}
Kaiming He, Xiangyu Zhang, Shaoqing Ren, and Jian Sun.
\newblock Deep residual learning for image recognition.
\newblock In {\em CVPR}, 2015.

\bibitem{Gate}
Xinrui Jiang, Ye Zhang, Qi Yang, Bin Deng, and Hongqiang Wang.
\newblock Millimeter-wave array radar-based human gait recognition using
  multi-channel three-dimensional convolutional neural network.
\newblock {\em Sensors}, 2020.

\bibitem{kaul_rss-net_2020}
Prannay Kaul, Daniele De~Martini, Matthew Gadd, and Paul Newman.
\newblock {RSS}-{Net}: {weakly}-{supervised} {multi}-{class} {semantic}
  {segmentation} with {FMCW} {radar}.
\newblock In {\em IV}, 2020.

\bibitem{mulran_kim_2020}
Giseop Kim, Yeong~Sang Park, Younghun Cho, Jinyong Jeong, and Ayoung Kim.
\newblock Mulran: Multimodal range dataset for urban place recognition.
\newblock In {\em ICRA}, 2020.

\bibitem{kingma_adam_2015}
Diederik~P. Kingma and Jimmy Ba.
\newblock Adam: {A} {method} for {stochastic} {optimization}.
\newblock In {\em ICLR}, 2015.

\bibitem{pointpillars}
Alex~H. Lang, Sourabh Vora, Holger Caesar, Lubing Zhou, Jiong Yang, and Oscar
  Beijbom.
\newblock Pointpillars: Fast encoders for object detection from point clouds.
\newblock In {\em CVPR}, 2019.

\bibitem{FPN}
Tsung{-}Yi Lin, Piotr Doll{\'{a}}r, Ross~B. Girshick, Kaiming He, Bharath
  Hariharan, and Serge~J. Belongie.
\newblock Feature pyramid networks for object detection.
\newblock In {\em CVPR}, 2017.

\bibitem{focal_loss}
Tsung{-}Yi Lin, Priya Goyal, Ross~B. Girshick, Kaiming He, and Piotr
  Doll{\'{a}}r.
\newblock Focal loss for dense object detection.
\newblock In {\em ICCV}, 2017.

\bibitem{lombacher_semantic_2017}
Jakob Lombacher, Kilian Laudt, Markus Hahn, Jurgen Dickmann, and Christian
  Wohler.
\newblock Semantic radar grids.
\newblock In {\em {IV}}, 2017.

\bibitem{major_vehicle_2019}
Bence Major, Daniel Fontijne, Amin Ansari, Ravi~Teja Sukhavasi, Radhika
  Gowaikar, Michael Hamilton, Sean Lee, Slawomir Grzechnik, and Sundar
  Subramanian.
\newblock Vehicle {detection} {with} {automotive} {radar} {using} {deep}
  {learning} on {range}-{azimuth}-{doppler} {tensors}.
\newblock In {\em ICCV Workshops}, 2019.

\bibitem{automotive_Meyer_2019}
Michael Meyer and Georg Kuschk.
\newblock Automotive radar dataset for deep learning based 3d object detection.
\newblock In {\em EuRAD}, 2019.

\bibitem{meyer_deep_2019}
Michael Meyer and Georg Kuschk.
\newblock Deep learning based 3d object detection for automotive radar and
  camera.
\newblock In {\em EuRAD}, 2019.

\bibitem{Aerial}
Casian Miron, Alexandru Pasarica, and Radu Timofte.
\newblock Efficient cnn architecture for multi-modal aerial view object
  classification.
\newblock In {\em CVPR Workshops}, 2021.

\bibitem{zendar}
Mohammadreza Mostajabi, Ching~Ming Wang, Darsh Ranjan, and Gilbert Hsyu.
\newblock High resolution radar dataset for semi-supervised learning of dynamic
  objects.
\newblock In {\em CVPR Workshop}, 2020.

\bibitem{ng_range-doppler_2020}
Weichong Ng, Guohua Wang, {Siddhartha}, Zhiping Lin, and Bhaskar~Jyoti Dutta.
\newblock Range-{Doppler} {detection} in {automotive} {radar} with {deep}
  {learning}.
\newblock In {\em {IJCNN}}, 2020.

\bibitem{nowruzi_polarnet_2021}
Farzan~Erlik Nowruzi, Dhanvin Kolhatkar, Prince Kapoor, Elnaz~Jahani Heravi,
  Fahed~Al Hassanat, Robert Lagani{\`{e}}re, Julien Rebut, and Waqas Malik.
\newblock Polarnet: Accelerated deep open space segmentation using automotive
  radar in polar domain.
\newblock In {\em VEHITS}, 2021.

\bibitem{ouaknine_multi-view_2021}
Arthur Ouaknine, Alasdair Newson, Patrick P{\'{e}}rez, Florence Tupin, and
  Julien Rebut.
\newblock Multi-view radar semantic segmentation.
\newblock In {\em ICCV}, 2021.

\bibitem{carrada_ouaknine_2021}
Arthur Ouaknine, Alasdair Newson, Julien Rebut, Florence Tupin, and Patrick
  Pérez.
\newblock Carrada dataset: Camera and automotive radar with range- angle-
  doppler annotations.
\newblock In {\em ICPR}, 2021.

\bibitem{palffy_cnn_2020}
Andras Palffy, Jiaao Dong, Julian F.~P. Kooij, and Dariu~M. Gavrila.
\newblock {CNN} based {road} {user} {detection} using the {3D} {radar} {cube}.
\newblock In {\em RAL}, 2020.

\bibitem{prophet_semantic_2020}
Robert Prophet, Anastasios Deligiannis, Juan-Carlos Fuentes-Michel, Ingo Weber,
  and Martin Vossiek.
\newblock Semantic {segmentation} on {3D} {occupancy} {grids} for {automotive}
  {radar}.
\newblock In {\em IEEE Access}, 2020.

\bibitem{prophet_semantic_2019}
Robert Prophet, Gang Li, Christian Sturm, and Martin Vossiek.
\newblock Semantic {segmentation} on {automotive} {radar} {maps}.
\newblock In {\em {IV}}, 2019.

\bibitem{scheiner_seeing_2020}
Nicolas Scheiner, Florian Kraus, Fangyin Wei, Buu Phan, Fahim Mannan, Nils
  Appenrodt, Werner Ritter, Jürgen Dickmann, Klaus Dietmayer, Bernhard Sick,
  and Felix Heide.
\newblock Seeing {around} {street} {corners}: {non}-{line}-of-{sight}
  {detection} and {tracking} {in}-the-{wild} {using} {Doppler} {radar}.
\newblock In {\em CVPR}, 2020.

\bibitem{radarscenes_Schumann_2021}
Ole Schumann, Markus Hahn, Nicolas Scheiner, Fabio Weishaupt, Julius~F. Tilly,
  J{\"{u}}rgen Dickmann, and Christian W{\"{o}}hler.
\newblock Radarscenes: {A} real-world radar point cloud data set for automotive
  applications.
\newblock In {\em ArXiv}, 2021.

\bibitem{schumann_scene_2020}
Ole Schumann, Jakob Lombacher, Markus Hahn, Christian Wohler, and Jurgen
  Dickmann.
\newblock Scene {understanding} {with} {automotive} {radar}.
\newblock In {\em IV}, 2020.

\bibitem{sheeny_radiate_2020}
Marcel Sheeny, Emanuele De~Pellegrin, Saptarshi Mukherjee, Alireza Ahrabian,
  Sen Wang, and Andrew Wallace.
\newblock {RADIATE}: {a} {radar} {dataset} for {automotive} {perception}.
\newblock In {\em ArXiv}, 2020.

\bibitem{sless_road_2019}
Liat Sless, Gilad Cohen, Bat~El Shlomo, and Shaul Oron.
\newblock Road {scene} {understanding} by {occupancy} {grid} {learning} from
  {sparse} {radar} {clusters} using {semantic} {segmentation}.
\newblock In {\em ICCV Workshops}, 2019.

\bibitem{wang_rodnet_2021}
Yizhou Wang, Zhongyu Jiang, Yudong Li, Jenq-Neng Hwang, Guanbin Xing, and Hui
  Liu.
\newblock {RODNet}: {a} {real}-{time} {radar} {object} {detection} {network}
  {cross}-{supervised} by {camera}-{radar} {fused} {object} {3D}
  {localization}.
\newblock In {\em SP}, 2021.

\bibitem{rethinking_whang_2021}
Yizhou Wang, Gaoang Wang, Hung{-}Min Hsu, Hui Liu, and Jenq{-}Neng Hwang.
\newblock Rethinking of radar's role: {A} camera-radar dataset and systematic
  annotator via coordinate alignment.
\newblock In {\em CVPR Workshop}, 2021.

\bibitem{pixor}
Bin Yang, Wenjie Luo, and Raquel Urtasun.
\newblock {PIXOR:} real-time 3d object detection from point clouds.
\newblock In {\em CVPR}, 2018.

\bibitem{raddet_Zhang_2021}
Ao Zhang, Farzan~Erlik Nowruzi, and Robert Lagani{\`{e}}re.
\newblock Raddet: Range-azimuth-doppler based radar object detection for
  dynamic road users.
\newblock In {\em CVR}, 2021.

\bibitem{zhang_object_2020}
Guoqiang Zhang, Haopeng Li, and Fabian Wenger.
\newblock Object {detection} and 3d {estimation} {via} an {FMCW} {radar}
  {using} a {fully} {convolutional} {network}.
\newblock In {\em {ICASSP}}, 2020.

\bibitem{PSPNet}
Hengshuang Zhao, Jianping Shi, Xiaojuan Qi, Xiaogang Wang, and Jiaya Jia.
\newblock Pyramid scene parsing network.
\newblock In {\em CVPR}, 2017.

\end{thebibliography}
}

\appendix

\section{Details of the \dataset~dataset}
\label{sec:append_dataset}

\paragnoskip{Sensor specifications.}
Central to the proposed \dataset~dataset, our high-definition radar is composed of $\Rx = 16$ receiving antennas and $\Tx = 12$ transmitting antennas, leading to $\Rx{\cdot}\Tx=192$ virtual antennas. This virtual-antenna array enables reaching a high azimuth angular resolution while estimating objects' elevation angles as well. 
As the radar signal is difficult to interpret by annotators and practitioners alike, a 16-layer automotive-grade laser scanner (LiDAR) and a 5 Mpix RGB camera are also provided. The camera is placed below the interior mirror behind the windshield while the radar and the LiDAR are installed in the middle of the front ventilation grid, one above the other. The three sensors have parallel horizontal lines of sight, pointing in the driving direction. Their extrinsic parameters are provided together with the dataset.
\dataset~also offers synchronized GPS and CAN traces which give access to the geo-referenced position of the vehicle as well as its driving information such as speed, steering wheel angle and yaw rate.
The sensors' specifications are detailed in Table \ref{tab:TableSensorsSpecs}.

\begin{figure}[h]
    \centering
    \includegraphics[width=0.7\columnwidth]{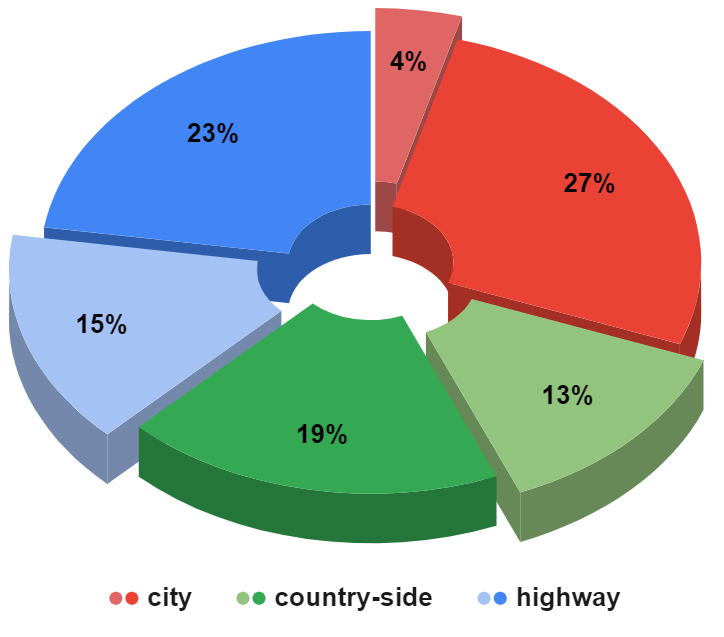}
    \vspace{-1em}\caption{\label{dataset}\textbf{Scene-type proportions  in \dataset}. The dataset contains 91 sequences in total, captured on city streets, highway or country-side roads, for a total of 25k synchronized frames (dark colors), out of which 8,252 are labelled (light colors).}
\end{figure}

\parag{\dataset~dataset.}
\dataset~contains 91 sequences of 1 to 4 minutes in duration, for a total of 2 hours. These sequences are categorized in \textit{highway}, \textit{country-side} and \textit{city} driving. The distribution of the sequences is indicated in Figure \ref{dataset}.
Each sequence contains raw sensor signals recorded with their native frame rate. A Python library is provided to read and synchronize the data together. There are approximately 25,000 frames with the three sensors synchronized, out of which 8,252 are labelled with a total of 9,550 vehicles.

\begin{table}[h]
\vspace{1em}\centering
\resizebox{\columnwidth}{!}{%
\begin{tabular}{clccc}
\toprule
 & & \textbf{HD Radar} &  \textbf{LiDAR} & \textbf{Camera}\\
\midrule
\multirow{3}{*}{\rotatebox{90}{FOV}} & Range   & 103\,m & 150\,m & --\\
& Azimuth   & $180^{\circ}$ & $133^{\circ}$ & $100^{\circ}$\\
& Elevation  & $12^{\circ}$ & $10^{\circ}$ & $75^{\circ}$\\
\midrule
\multirow{4}{*}{\rotatebox{90}{resolution}} & Range & 0.2\,m & 0.1\,m & \\
& Azimuth & $0.1^{\circ}$& $0.125$-$0.25^{\circ}$ & $2592$\,px\\
& Elevation & $1^{\circ}$ & $0.6^{\circ}$ & $1944$\,px \\
& Velocity & $0.1\,\text{m}{\cdot}\text{s}^{-1}$ & -- & -- \\
\midrule
\multicolumn{2}{l}{Frame rate} & 5\,fps & 25\,fps & 30\,fps \\
\multicolumn{2}{l}{Height above ground} &  80\,cm & 42\,cm & 145\,cm \\
\bottomrule
\end{tabular}%
}
\caption{\label{tab:TableSensorsSpecs}\textbf{Specification of the  \dataset's sensor suite}. The main characteristics of the HD radar, the LiDAR and the camera are reported. Their synchronized signals are complemented by GPS and CAN information.}
\end{table}

\section{Ablation study of the MIMO pre-encoder}
\label{sec:append_ablation}
\begin{figure}[t]
    \centering
    \includegraphics[width=1.0\columnwidth]{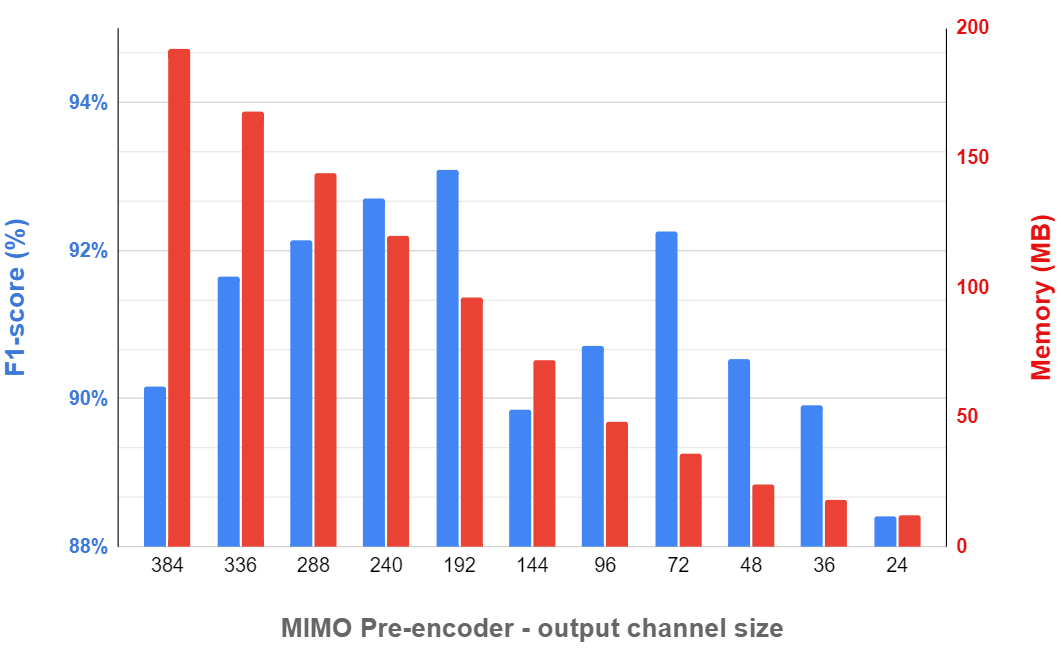}
    \vspace{-2em}\caption{\label{ablation}\textbf{MIMO pre-encoder ablation}. Influence of the number of output channels of the pre-encoder on the memory footprint and the performance of the detection head.}
\end{figure}

The role of the MIMO pre-encoder is to de-interleave the range-Doppler spectrums and to transform them into a representation that is compact and still allows, through learning, the prediction of azimuth angles along with other information on reflectors. 
The input of the MIMO pre-encoder is composed of the $\Rx = 16$ range-Doppler spectrums \textit{in complex numbers}, one for each Rx.  The real and imaginary parts are stacked, yielding an input tensor of total size $B_\text{R}{\times}B_\text{D}{\times}2\Rx$, \ie, $512{\times}256{\times}32$.   
The ablation study consists in evaluating the performance of \method's detection head while reducing the number of feature channels that the MIMO pre-encoder outputs. 
The maximum number of output channels is the number of virtual antennas with a complex signal (real and imaginary parts), \ie, 
$\Tx{\cdot}2{\cdot}\Rx  = 384$. We vary the number of output channels from a minimum of 24 to this maximum value and compute the detection performance on the validation set.
The results of this ablation study are reported in Figure \ref{ablation}. We measure the detection performance with the F1-score, classically defined as  $\text{F1-score} = \frac{\text{AP}\cdot\text{AR}}{\text{AP} + \text{AR}}$, which aggregates in a single metric both the Average Precision (AP) and the Average Recall (AR).
We observe that the best performance is reached with 192 output channels, hence half of the maximum output size. 
This compressed output is the one that captures at best the range and azimuth information from the inputs range-Doppler spectrums toward the detection and segmentation tasks. 

\section{Radar versus LiDAR}
\label{sec:append_lidar}

\dataset~dataset was designed to collect information from several sensor technologies. For safety-critical systems, such as self-driving vehicles, we believe that redundancy at various levels of the system, starting at the sensing layer, is key to guarantee safe operations.  
In a complete automated driving system, the combination of radars together with cameras and LiDARs will improve the overall robustness. Indeed, LiDARs provide, even at night, accurate 3D localization of objects in distance and angle, while cameras give access to a wealth of semantic and geometric information about the scene when light is sufficient. However, these two types of sensors suffer from bad weather conditions that can degrade quite significantly their performances. Radars are more robust to adverse weather conditions, provide accurate distance estimates together with the velocity of the objects, and are especially well suited to the cost and size constraints of automotive applications. 

For reference, we report in Table \ref{tab:Perf_objects2} the performance on \dataset~obtained by the imaging radar (with FFT-RadNet) and by the LiDAR sensor (with Pixor), respectively. The former obtains similar performances in AP and lower, though still good, performances in AR compared to the latter. This is already a remarkable result, owing to the practical advantages of the radar technology, which we reminded above. 
In addition, this difference of performances might be explained by the way \dataset~dataset was created. The ground truth is 
obtained semi-automatically based on 2D detection/segmentation from the camera fused with the 3D information of LiDAR. 
The evaluation might thus be favorably biased toward processing LiDAR inputs. 

\begin{table}[h]
\vspace{1em}\centering
\begin{tabular}{l | c| c| c c}
\toprule
Sensor & Model & AP($\%$) $\uparrow$ & AR($\%$) $\uparrow$\\ 
\toprule
    LiDAR & Pixor & 98.55 & 90.42\\
    Radar & FFT-RadNet & 96.84 & 82.18\\
 \bottomrule

\end{tabular}
\caption{\label{tab:Perf_objects2}\textbf{Vehicle detection with HD radar alone and LiDAR alone}.
Performance in average precision (AP) and average recall (AR) on RADIal Test split. FFT-RadNet takes range-Doppler spectrum as input, and Pixor the LiDAR point cloud. 
}
\end{table}

\begin{figure*}[t]
\renewcommand{\captionfont}{\small}
\centering
\scriptsize
  \begin{tabular}{c c | c c}
    
    FFT-RadNet & PIXOR LiDAR & FFT-RadNet & PIXOR LiDAR\\
    \includegraphics[width=4cm]{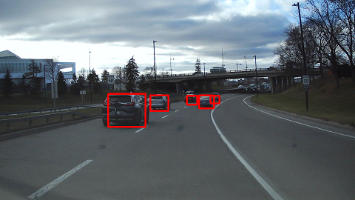} & 
    \includegraphics[width=4cm]{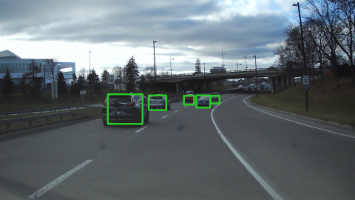} &
    
    \includegraphics[width=4cm]{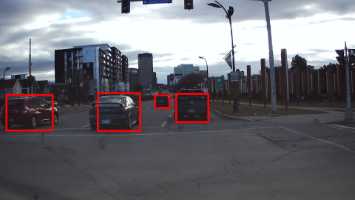} & 
    \includegraphics[width=4cm]{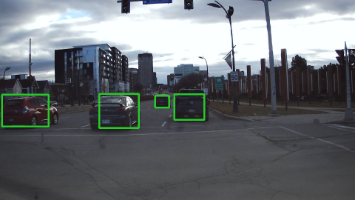} \\
    
    \includegraphics[width=4cm]{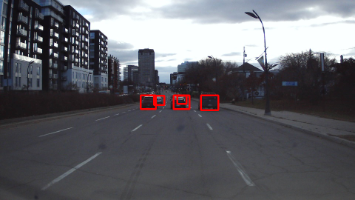} & 
    \includegraphics[width=4cm]{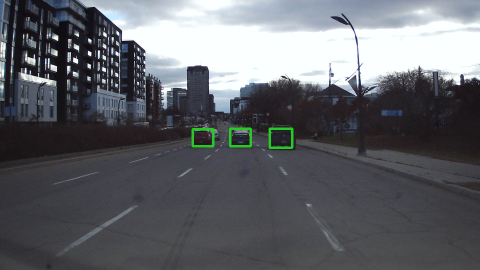} &
    
    \includegraphics[width=4cm]{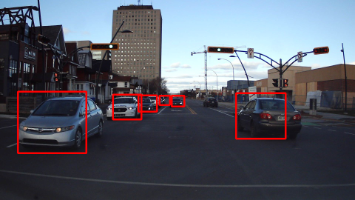} & 
    \includegraphics[width=4cm]{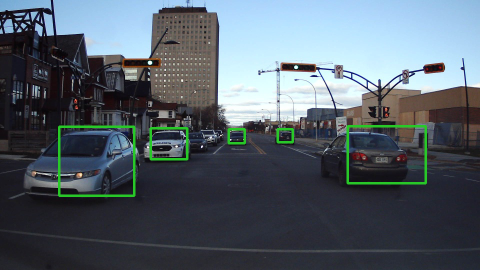} \\
    
    \includegraphics[width=4cm]{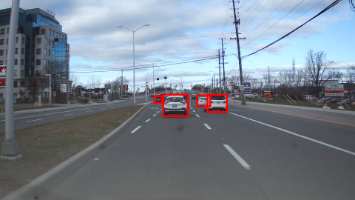} & 
    \includegraphics[width=4cm]{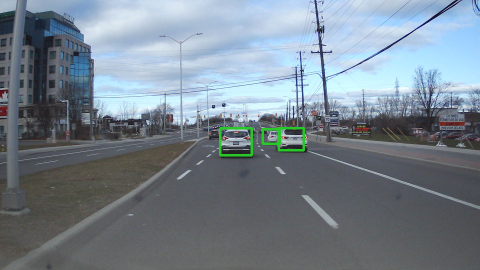} &
    
    \includegraphics[width=4cm]{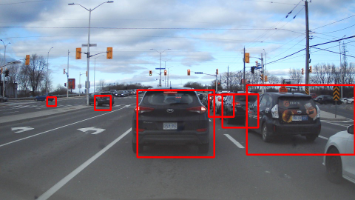} & 
    \includegraphics[width=4cm]{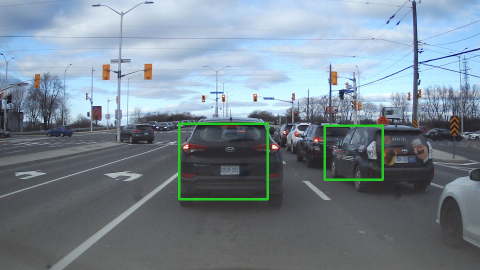} \\
    
    \includegraphics[width=4cm]{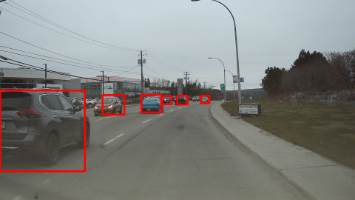} & 
    \includegraphics[width=4cm]{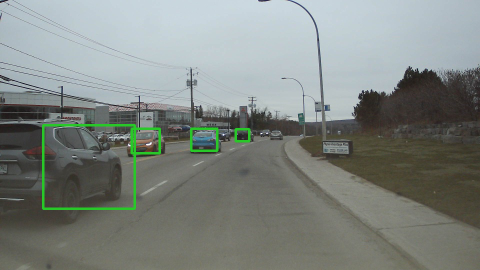} &
    
    \includegraphics[width=4cm]{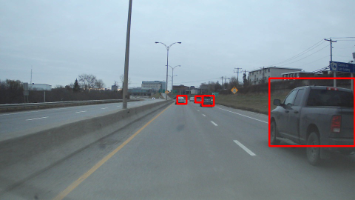} & 
    \includegraphics[width=4cm]{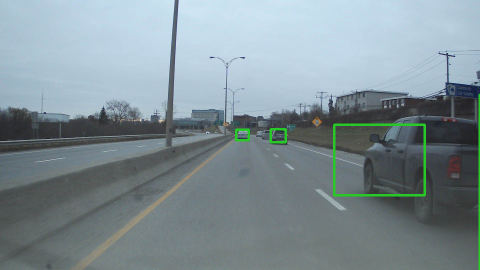} \\
    
    \includegraphics[width=4cm]{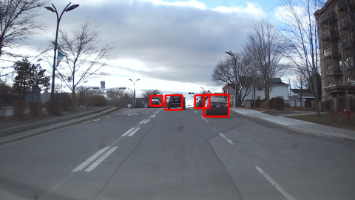} & 
    \includegraphics[width=4cm]{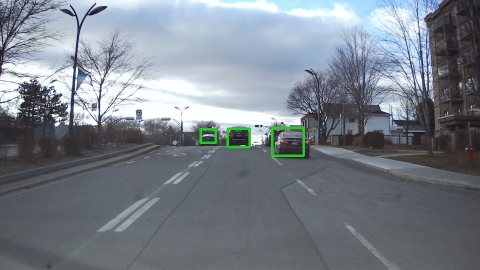} &
    
    \includegraphics[width=4cm]{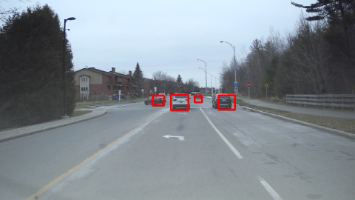} & 
    \includegraphics[width=4cm]{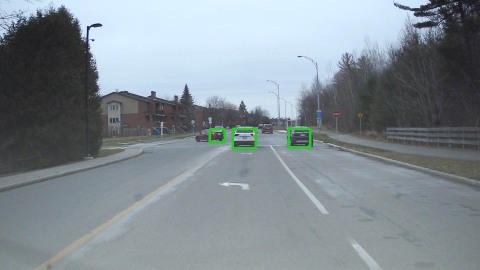} \\
    
    \includegraphics[width=4cm]{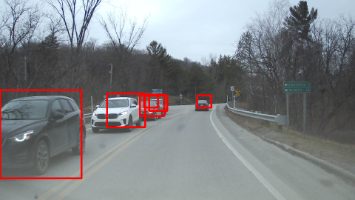} & 
    \includegraphics[width=4cm]{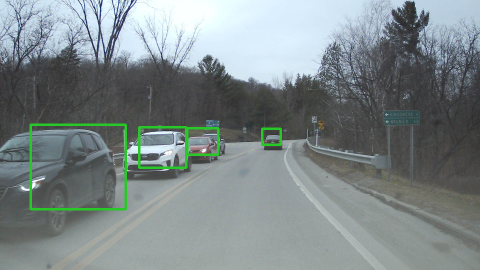} &
    
    \includegraphics[width=4cm]{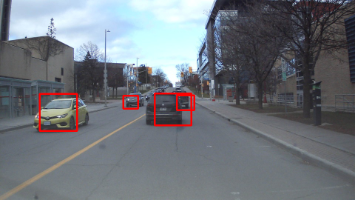} & 
    \includegraphics[width=4cm]{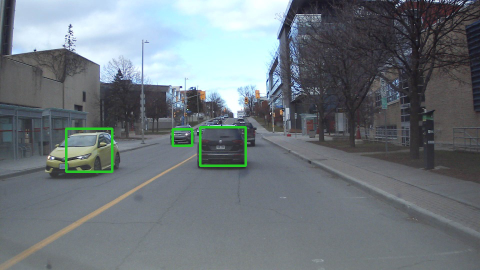} \\
    
    \includegraphics[width=4cm]{Images/Qualitative/CVPR_dense_011009.png} & 
    \includegraphics[width=4cm]{Images/Qualitative/CVPR_dense_SCALA011009.png} &
    
    \includegraphics[width=4cm]{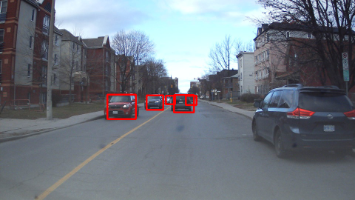} & 
    \includegraphics[width=4cm]{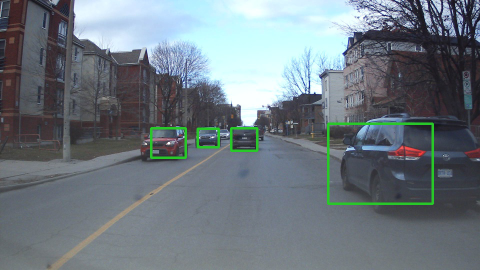} \\
    
    \includegraphics[width=4cm]{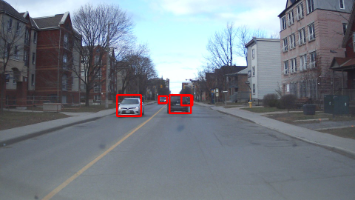} & 
    \includegraphics[width=4cm]{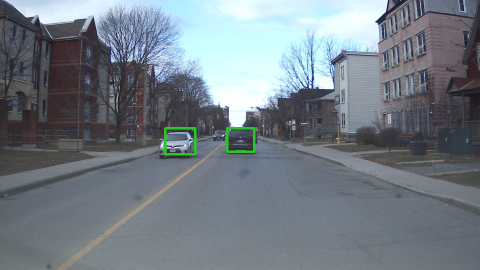} &
    \includegraphics[width=4cm]{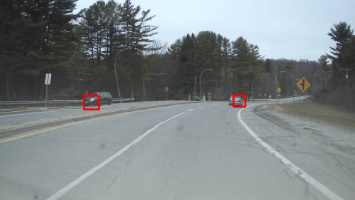} & 
    \includegraphics[width=4cm]{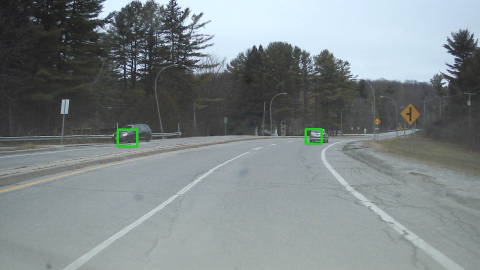} \\
    
    \includegraphics[width=4cm]{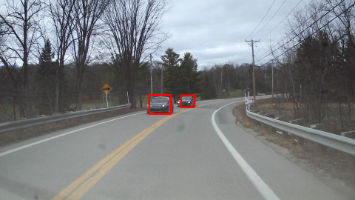} & 
    \includegraphics[width=4cm]{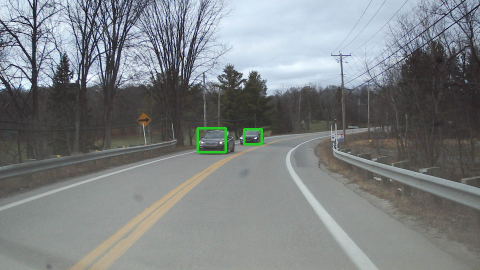} &
    \includegraphics[width=4cm]{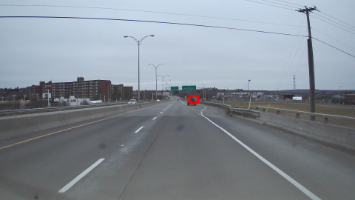} & 
    \includegraphics[width=4cm]{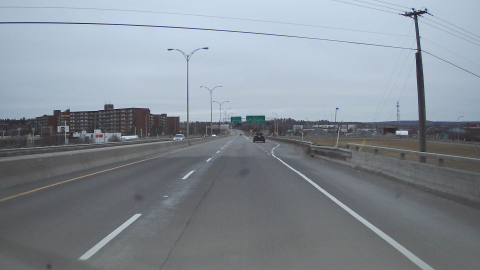} \\
    
  \end{tabular}
  \caption{
  \label{fig:qualitative2}
    \textbf{Examples of vehicle detection in complex scenes using either HD radar or LiDAR}.
    Comparison between Pixor trained with LiDAR point clouds (`PIXOR LiDAR' columns, green boxes) and our proposed FFT-RadNet requiring only range-Doppler as input (`FFT-RadNet', red boxes). Note that radar detections are not limited to the first row of vehicles but can see up to the second one. Also, FFT-RadNet provides vehicles'  relative speed through Doppler measurements.}
\end{figure*}

Due to the nature of the annotation pipeline, and to the radar multi-path reflections, many sequences of complex scenes in urban or dense environments, which are present in \dataset, were not annotated. 
In Figure \ref{fig:qualitative2}, we qualitatively compare vehicle detection in such complex scenes when using either the HD radar modality or the LiDAR one. 
We observe that the HR radar, equipped with FFT-RadNet, detects vehicles in complex situations, including beyond the first row of vehicles where neither the camera nor the LiDAR performs well.

\end{document}